\documentclass[10pt,twocolumn,letterpaper]{article}

\usepackage{iccv}
\usepackage{times}
\usepackage{epsfig}
\usepackage{graphicx}
\usepackage{amsmath}
\usepackage{amssymb}
\usepackage{array}
\usepackage{booktabs}
\usepackage{pifont}
\usepackage{float}
\usepackage{cuted}
\usepackage{capt-of}
\usepackage{subcaption}
\usepackage{enumitem}

\usepackage[pagebackref=true,breaklinks=true,letterpaper=true,colorlinks,bookmarks=false]{hyperref}

\iccvfinalcopy 

\def\iccvPaperID{1832} 

\ificcvfinal\fi

\begin{document}
	

\newif\ifdraft
\drafttrue

\ifdraft
\newcommand{\PF}[1]{{\color{red}{\bf PF: #1}}}
\newcommand{\pf}[1]{{\color{red} #1}}
\newcommand{\IK}[1]{{\color{blue}{\bf IK: #1}}}
\newcommand{\ik}[1]{{\color{blue} #1}}
\newcommand{\HR}[1]{{\color{magenta}{\bf hr: #1}}}
\newcommand{\hr}[1]{{\color{magenta} #1}}
\newcommand{\VC}[1]{{\color{blue}{\bf vc: #1}}}
\newcommand{\vc}[1]{{\color{blue} #1}}
\newcommand{\ms}[1]{{\color{green}{#1}}}
\newcommand{\MS}[1]{{\color{green}{\bf ms: #1}}}
\newcommand{\JS}[1]{{\color{cyan}{\bf js: #1}}}
\newcommand{\NEW}[1]{{\color{red}{#1}}}

\else
\newcommand{\PF}[1]{{\color{red}{}}}
\newcommand{\pf}[1]{ #1 }
\newcommand{\HR}[1]{{\color{blue}{}}}
\newcommand{\hr}[1]{#1}%
\newcommand{\VC}[1]{{\color{green}{}}}
\newcommand{\ms}[1]{ #1 }
\newcommand{\MS}[1]{{\color{green}{}}}
\newcommand{\NEW}[1]{#1}
\fi

\newcommand{\comment}[1]{}

\newcommand{\va}{\mathbf{a}}
\newcommand{\vb}{\mathbf{b}}
\newcommand{\vcc}{\mathbf{c}}
\newcommand{\vd}{\mathbf{d}}
\newcommand{\ve}{\mathbf{e}}
\newcommand{\vf}{\mathbf{f}}
\newcommand{\vg}{\mathbf{g}}
\newcommand{\vh}{\mathbf{h}}
\newcommand{\vi}{\mathbf{i}}
\newcommand{\vj}{\mathbf{j}}
\newcommand{\vk}{\mathbf{k}}
\newcommand{\vl}{\mathbf{l}}
\newcommand{\vm}{\mathbf{m}}
\newcommand{\vn}{\mathbf{n}}
\newcommand{\vo}{\mathbf{o}}
\newcommand{\vp}{\mathbf{p}}
\newcommand{\vq}{\mathbf{q}}
\newcommand{\vr}{\mathbf{r}}
\newcommand{\vt}{\mathbf{t}}
\newcommand{\vu}{\mathbf{u}}
\newcommand{\vv}{\mathbf{v}}
\newcommand{\vw}{\mathbf{w}}
\newcommand{\vx}{\mathbf{x}}
\newcommand{\vy}{\mathbf{y}}
\newcommand{\vz}{\mathbf{z}}

\newcommand{\mA}{\mathbf{A}}
\newcommand{\mB}{\mathbf{B}}
\newcommand{\mC}{\mathbf{C}}
\newcommand{\mD}{\mathbf{D}}
\newcommand{\mE}{\mathbf{E}}
\newcommand{\mF}{\mathbf{F}}
\newcommand{\mG}{\mathbf{G}}
\newcommand{\mH}{\mathbf{H}}
\newcommand{\mI}{\mathbf{I}}
\newcommand{\mJ}{\mathbf{J}}
\newcommand{\mK}{\mathbf{K}}
\newcommand{\mL}{\mathbf{L}}
\newcommand{\mM}{\mathbf{M}}
\newcommand{\mN}{\mathbf{N}}
\newcommand{\mO}{\mathbf{O}}
\newcommand{\mP}{\mathbf{P}}
\newcommand{\mQ}{\mathbf{Q}}
\newcommand{\mR}{\mathbf{R}}
\newcommand{\mS}{\mathbf{S}}
\newcommand{\mT}{\mathbf{T}}
\newcommand{\mU}{\mathbf{U}}
\newcommand{\mV}{\mathbf{V}}
\newcommand{\mW}{\mathbf{W}}
\newcommand{\mX}{\mathbf{X}}
\newcommand{\mY}{\mathbf{Y}}
\newcommand{\mZ}{\mathbf{Z}}

\newcommand{\cA}{\mathcal A}
\newcommand{\cB}{\mathcal B}
\newcommand{\cC}{\mathcal C}
\newcommand{\cD}{\mathcal D}
\newcommand{\cE}{\mathcal E}
\newcommand{\cF}{\mathcal F}
\newcommand{\cG}{\mathcal G}
\newcommand{\cH}{\mathcal H}
\newcommand{\cI}{\mathcal I}
\newcommand{\cJ}{\mathcal J}
\newcommand{\cK}{\mathcal K}
\newcommand{\cL}{\mathcal L}
\newcommand{\cM}{\mathcal M}
\newcommand{\cN}{\mathcal N}
\newcommand{\cO}{\mathcal O}
\newcommand{\cP}{\mathcal P}
\newcommand{\cQ}{\mathcal Q}
\newcommand{\cR}{\mathcal R}
\newcommand{\cS}{\mathcal S}
\newcommand{\cT}{\mathcal T}
\newcommand{\cU}{\mathcal U}
\newcommand{\cV}{\mathcal V}
\newcommand{\cW}{\mathcal W}
\newcommand{\cX}{\mathcal X}
\newcommand{\cY}{\mathcal Y}
\newcommand{\cZ}{\mathcal Z}

\newcommand*\rot{\rotatebox{90}}
\newcommand*\OK{\ding{51}}

\newcommand{\TODO}[1]{\textcolor{cyan}{#1}}
\definecolor{gray}{RGB}{127,127,127}
\newcommand{\optional}[1]{\textcolor{gray}{#1}}

\newcommand{\ST}{\mathcal{T}}
\newcommand{\SST}{\mathcal{T}_S}

\newcommand{\R}{\mathbb{R}}
\newcommand{\Seg}{\mathbf{S}} 
\newcommand{\Latent}{\mathbf{L}}
\newcommand{\LatentG}{\Latent^{\text{3D}}} 
\newcommand{\LatentA}{\Latent^\text{app}} 
\newcommand{\LatentBG}{\mB} 

\newcommand{\loss}{L}
\newcommand{\objFG}{O}
\newcommand{\objBG}{G}

\newcommand{\norm}[1]{\left\lVert#1\right\rVert}
\newcommand{\argmin}{\operatornamewithlimits{argmin}}
\newcommand{\erf}{\operatornamewithlimits{erf}}

\newcommand{\Var}{\operatornamewithlimits{Var}}

\newcommand{\parag}[1]{\vspace{-4mm}\paragraph{#1}}
\newcommand{\sparag}[1]{\vspace{-3mm}\subparagraph{#1}}

\newcommand{\handheld}[0]{{\bf Handheld190k}}
\newcommand{\ski}[0]{{\bf Ski-PTZ}}
\newcommand{\human}[0]{{\bf H36M}}

\newcommand{\ours}[0]{{\bf Ours}}
\newcommand{\direct}[0]{{\bf Resnet}}
\newcommand{\LCR}[0]{{\bf LCR}}
\newcommand{\ECCV}[0]{{\bf NVS-encoder}}
\newcommand{\CVPR}[0]{{\bf Multiview}}
\newcommand{\auto}[0]{{\bf Auto-encoder}}

\title{Human Detection and Segmentation via Multi-view Consensus}

\author{
	Isinsu Katircioglu$^1$ \thanks{Work supported in part by the Swiss National Science Foundation.}
	\and
	Helge Rhodin$^2$
	\and
	J\"{o}rg Sp\"{o}rri$^3$
	\and
	Mathieu Salzmann$^{1,4}$
	\and
	Pascal Fua$^1$   
	\and
	$^1$EPFL, Lausanne, Switzerland\quad
	$^2$UBC, Vancouver, Canada\\
	$^3$Balgrist University Hospital, Zurich, Switzerland\quad
	$^4$ClearSpace SA, Lausanne, Switzerland\\
	{\tt\small \{firstname.lastname\}@epfl.ch, rhodin@cs.ubc.ca, joerg.spoerri@balgrist.ch}
}


\maketitle
\thispagestyle{empty}


\begin{strip}
\vspace{-11mm}
 \centering
  \begin{tabular}{@{}ccccc@{}}
\includegraphics[width=0.19\textwidth]{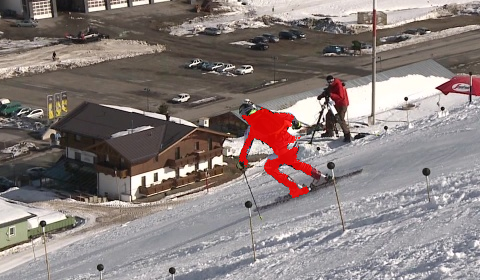} &\hspace{-3mm}
\includegraphics[width=0.19\textwidth]{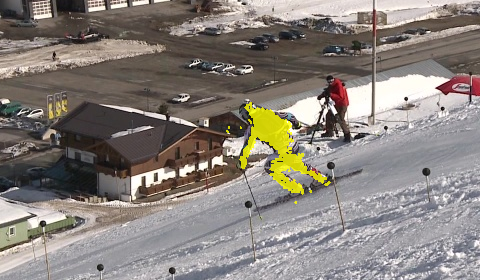} &\hspace{-3mm}
  \includegraphics[width=0.19\textwidth]{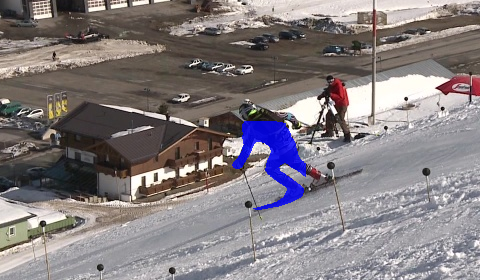} &\hspace{-4mm}
  \includegraphics[width=0.19\textwidth]{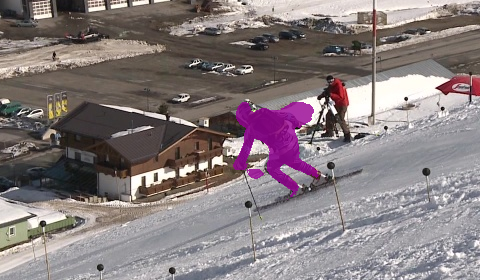} &\hspace{-4mm}
  \includegraphics[width=0.19\textwidth]{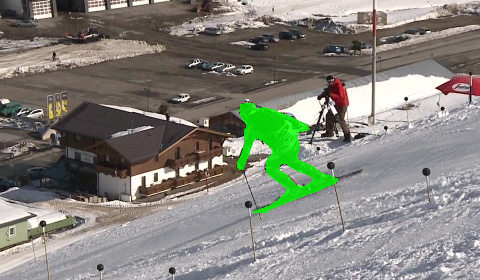} 
   \\
\includegraphics[width=0.19\textwidth]{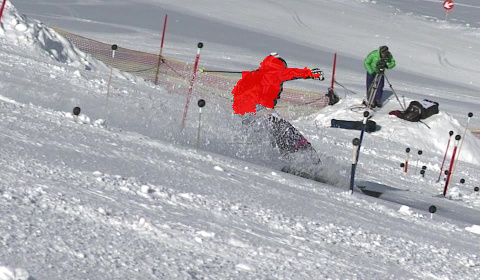} &\hspace{-3mm}
\includegraphics[width=0.19\textwidth]{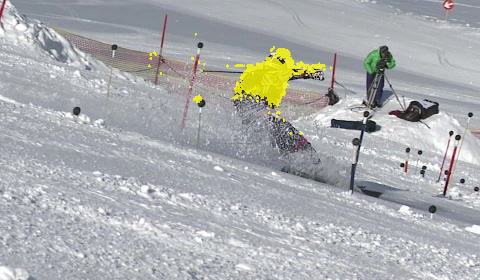} &\hspace{-3mm}
\includegraphics[width=0.19\textwidth]{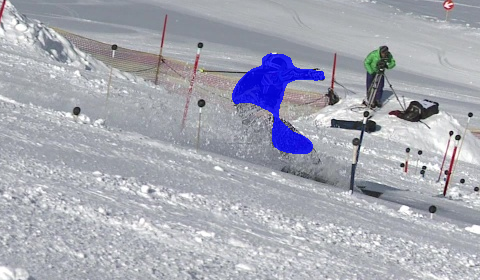} &\hspace{-4mm}
\includegraphics[width=0.19\textwidth]{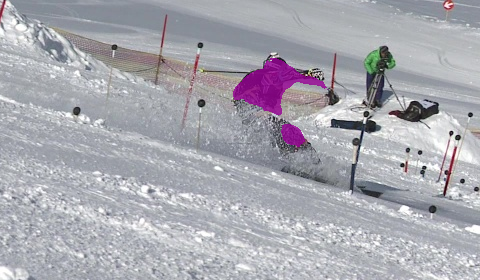} &\hspace{-4mm}
\includegraphics[width=0.19\textwidth]{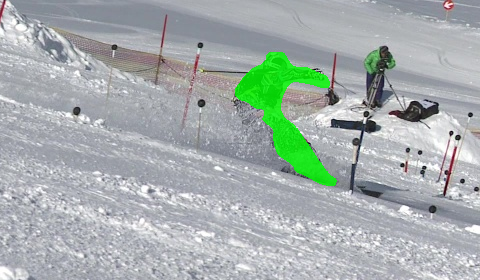} 
\\

\includegraphics[width=0.19\textwidth]{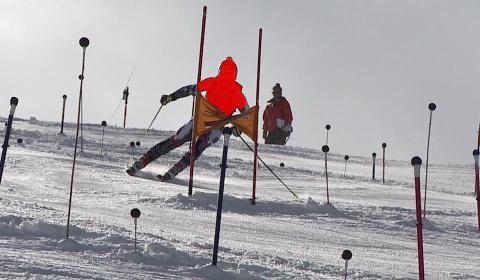} &\hspace{-3mm}
\includegraphics[width=0.19\textwidth]{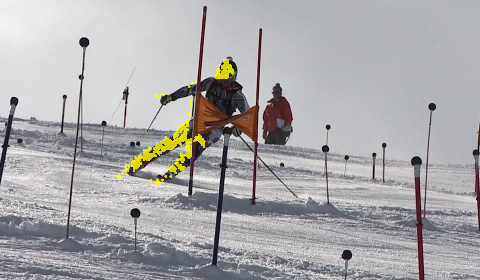} &\hspace{-3mm}
\includegraphics[width=0.19\textwidth]{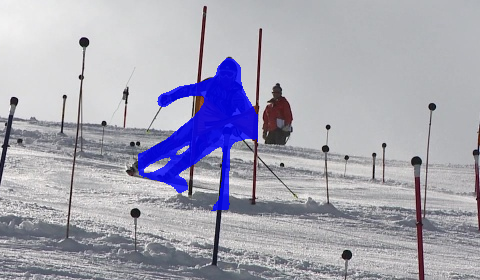} &\hspace{-4mm}
\includegraphics[width=0.19\textwidth]{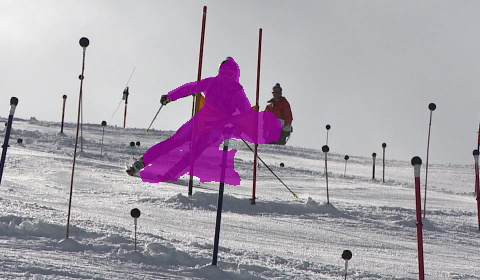} &\hspace{-4mm}
\includegraphics[width=0.19\textwidth]{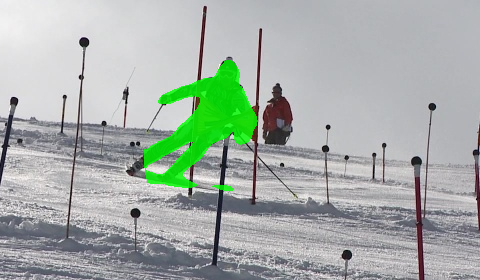} 
\\
    {\small {Koh et al.}~\cite{Koh17b} } & {\small {Yang et al.}~\cite{Yang19c} } & {\small {Katircioglu et al.}~\cite{Katircioglu20}}  &  {\small {Rhodin et al.}~\cite{Rhodin19a} + ~\cite{Katircioglu20} } &  {\small Ours }
  \end{tabular}
\vspace{-3mm}
\captionof{figure}{Leveraging multi-view consistency at training time to segment the salient object from {\it single} images at inference time and to outperform baselines exploiting temporal consistency~\cite{Koh17b}, optical flow~\cite{Yang19c,Katircioglu20} and novel view synthesis~\cite{Rhodin19a}.}
 \label{fig:teaser}
 
\end{strip}


\begin{abstract} 

\vspace{-4mm}
Self-supervised detection and segmentation of foreground objects aims for accuracy without annotated training data. However, existing approaches predominantly rely on restrictive assumptions on appearance and motion.

\vspace{-0mm}

For scenes with dynamic activities and camera motion,  we propose a multi-camera framework in which geometric constraints are embedded in the form of multi-view consistency during training via coarse 3D localization in a voxel grid and fine-grained offset regression. In this manner, we learn a joint distribution of proposals over multiple views. At inference time, our method operates on single RGB images.
We outperform state-of-the-art techniques  both on images that visually depart from those of standard benchmarks and on those of the classical Human3.6M dataset.

\end{abstract}
\vspace{-4mm}

\section{Introduction}

Robust detection and segmentation of moving people can now be achieved reliably in scenarios for which large amounts of annotated data are available. However, for less common activities, such as skiing, it remains challenging, because the required training databases do not exist. Self-supervised approaches~\cite{Eslami16,Koh17b,Bielski19,Chen19a,Crawford19,Croitoru19,Rhodin19a,Yang19c,Lin20a,Benny20,Lu20} promise to address this problem. However, most of them depend on strong constraints,  such as the target objects being seen against a static background, or rely on object localization and object-boundary detection networks pre-trained with supervision, which limits their applicability.


In this paper, we propose to remove these limitations by using a multi-camera setup for training purposes and explicitly encoding the 3D geometry of the scene. At inference time, our trained network can then handle single images and outperforms earlier techniques, as shown in Fig.~\ref{fig:teaser}. Our algorithm can be applied to any object as long as the two assumptions from~\cite{Leordeanu20} hold: foreground and background are distinguishable by color or texture; every part of the background must be visible more often than not.

Using several cameras complicates data acquisition but only in a limited way because both  synchronization and calibration are well understood tasks for which off-the-shelf solutions exist. In practice, for static cameras, this has to be dealt with only once before a filming session using well-known techniques~\cite{Hartley00,Faugeras01} and requires far less effort than manually annotating images. For moving cameras, SLAM methods are now robust enough to perform the calibration automatically and fast in the wild~\cite{Zou13,Wang18g}. Hence, there are many applications in which training with multiple cameras makes perfect sense, especially those with unusual activities for which large training databases are not available.

\begin{figure}
	\begin{center}
		\includegraphics[width=\linewidth]{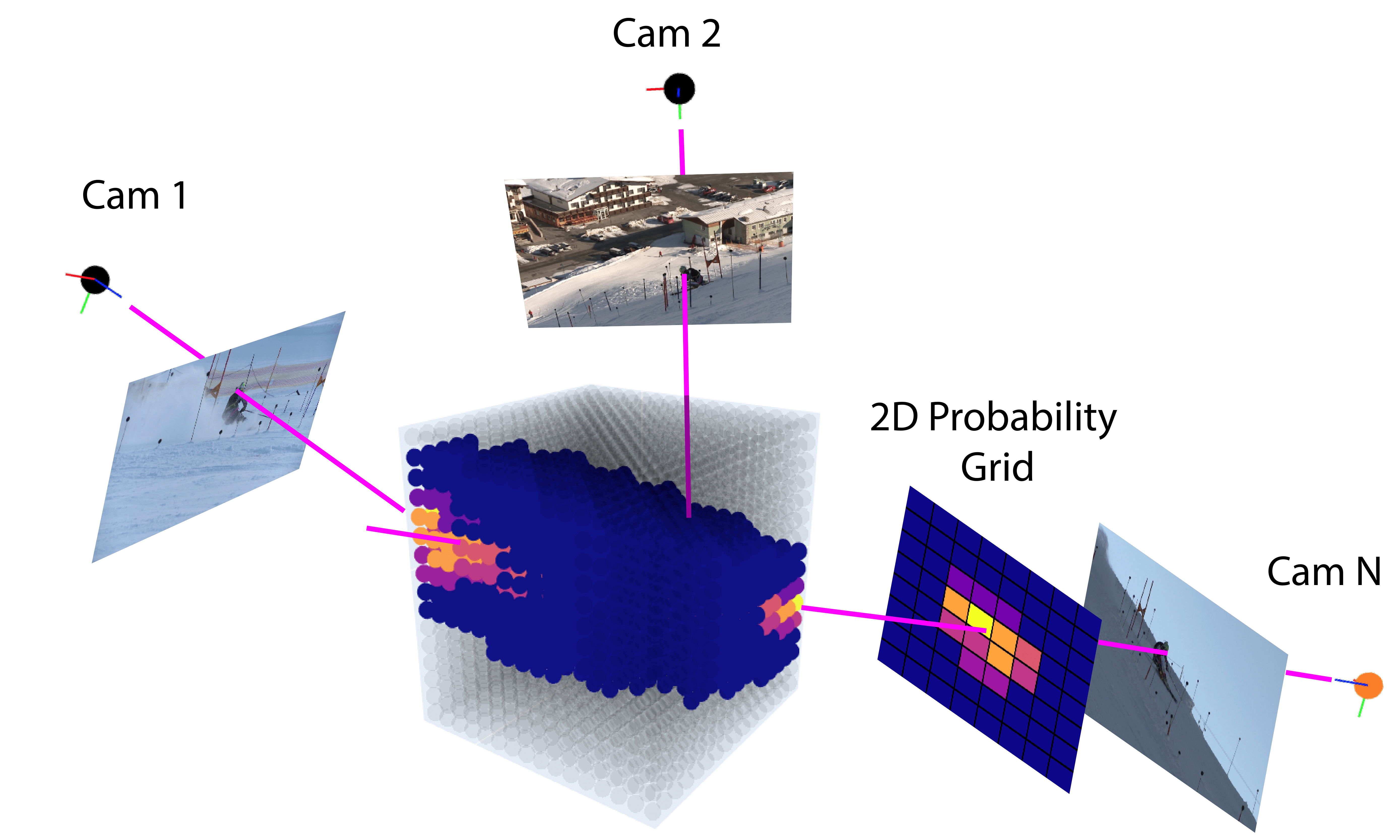}
	\end{center}
	\caption{\label{fig:Grid3D}
		 {\bf 3D Proposal Grid.} The consensus between individual views is found on a 3D voxel grid (black) as a combination of 2D probabilities projected on the voxels (rainbow colors). Once a coarse grid location is found, a fine offset is found via offset prediction and 3D triangulation (purple lines).
	}
\vspace{-4mm}
\end{figure}

To leverage multi-view training data as weak supervision, we introduce the object proposal strategy depicted by Fig.~\ref{fig:Grid3D}. Candidate 2D bounding boxes are produced by a network that can be trained in an unsupervised fashion. They are used to vote into a 3D proposal grid, and multi-view geometry constraints are then imposed to align proposals from different views in a differentiable manner. To train the resulting network, we sample a 3D proposal, deconstruct and reconstruct the image in each view using the corresponding 2D bounding box, and compare the resulting resynthesized images to the original ones.


While our self-supervised learning strategy leverages multiple views during training, the resulting model can be used for detection and segmentation in monocular images acquired by moving cameras and featuring unknown backgrounds. Our contributions can be summarized as follows.
\begin{itemize}
\item  We introduce a self-supervised  end-to-end trainable object detection and segmentation approach that explicitly leverages 3D multi-view geometry as weak supervision during training.
\item  It comprises a 3D object proposal framework that enables to enforce prediction consistency across views without having to introduce additional loss terms.
\end{itemize}

 To show that our approach can handle unusual activities and fast motion, we demonstrate it on the skiing dataset depicted by Fig.~\ref{fig:teaser}, captured by moving cameras, on a small dataset acquired using hand-held cameras, as well as on the more standard H36M dataset~\cite{Ionescu14a} acquired using fixed cameras. Note that our multi-view supervision differs from weak supervision in video object segmentation as it does not require any segmentation annotation. Hence, our method relates to self-supervised approaches. We show that the proposed multi-view training increases single-image accuracy performance at inference time, which allows us to outperform state-of-the-art single-view~\cite{Koh17b,Yang19c,Croitoru19,Lu20,Katircioglu20} and multi-view~\cite{Rhodin19a} approaches. Our code is publicly available at \url{https://github.com/isinsukatircioglu/mvc}.


\comment{


Most conventional approaches cast this problem as a multi-view reconstruction task to analyze the 3D spatial relationship between cameras and scene geometry. However, this is error-prone when cameras move quickly, when the scene offers only limited features, and when each camera only sees a small region with limited overlap. 
To deliver the best of both worlds, we combine a self-supervised detection and segmentation approach with a joint reasoning over different views that does not require explicit feature correspondences across time or views.

\HR{First talk about what you do, before you put it in context to what is out there.}
We propose a three-stage approach. First, coarse localization on a 3D voxel space that robustly determines the rough person position from individual proposals of each view via consensus voting. This is followed by the offset prediction stage that accurately triangulates the person position and scale as an offset to the output of the first stage. The third stage operates on a local crop around the object to predict the pixel-accurate segmentation mask. The first difficulty lies in setting up self-supervised objective.

We use a reconstruction objective that re-composes the input image from an estimate of foreground region and background image with the foreground inpainted. 
This is similar in spirit to a novel view synthesis based approach~\cite{Rhodin19a} that decomposes a single RGB image into foreground and background regions by exploiting the self-supervision coming from multi-view data and enforcing consistency among the results generated from different views. Given multiple views of the same image, the content of the predicted bounding box around that object in one view can be used to resynthesize the same object in another view. During training it uses a pair of views and at inference time it takes as input a single RGB image to detect and segment the foreground object. However, while doing this it assumes a known background and therefore can only work in a static camera setting, while our approach can handle moving cameras.
We therefore follow the observation from \cite{PAMI} that in most images the background forms a consistent, natural scene while the foreground is dynamic. Therefore, the appearance of any background patch can be predicted from its surroundings but the dynamic object is independent and cannot be inpainted. \cite{PAMI} therefore located the foreground object as that image region that is hardest to inpaint. It uses a 2D proposal network similar to Yolo~\cite{}, where the image is partitioned into a coarse 2D grid of occupancy probabilities and 2D offsets for precise localization.
However, this objective is error-prone as there can be inconsistent background regions. 

To this end, we reason in 3D, where false positives in 2D can be eliminated through multi-view analysis. This requires several advances.
First, we reason probabilistically by defining a joint probability over the 3D grid and marginalizing over the depth dimension to relate 3D voxels to 2D detections. 
Moreover, we combine the offset predictions from the selected 2D cells that correspond to the most likely 3D cell using a least squares formulation.
Finally, the consolidated object position and scale is re-projected to each view forming a crop window that is passed to an inpainting and segmentation network to separate foreground and background.

The difficulty is now to train this three-stage pipeline in an end-to-end fashion on the self-supervised image reconstruction objective.
Our core contribution are threefold. 1) Deriving a Monte Carlo sampling strategy that makes the 3D grid voting differentiable. 2) a bounding cylinder formulation that is scale invariant avoiding scale-depth ambiguities in monocular reconstruction, and 3) a differentiable triangulation module.

We show that....
}
\section{Related Work}
\label{sec:related}

Salient object detection and segmentation is a long-standing problem in computer vision. In this section, we review the monocular and multi-view approaches that have been proposed to solve this task. 

\parag{Single View Approaches.}

Most salient object detection and segmentation algorithms are fully-supervised~\cite{Cheng17a,He17a,Redmon16,Song18,Bhat20,Li20g,Lu20c,Seo20,Seong20,Yang20e} and require large annotated datasets containing pairs of images and labels. Our goal is to train a purely self-supervised method without either segmentation or object bounding box annotations. Note that this differs from the so-called \emph{unsupervised object segmentation} methods that leverage either domain-specific annotations during training but not at test time~\cite{Perazzi15,Hu18b,Jain17,Li18g,Li18j,Lu19,Wang19d,Yang19b,Wang19i,Zhang20c,Zhen20}, or the label of the first frame at inference time~\cite{Wang19g}. We focus our discussion on self- and  weakly-supervised methods with regard to the type of training data used.

As conventional methods relying on hand-crafted features, recent methods train deep neural networks for object detection~\cite{Jain17,Wei17}, optical flow estimation~\cite{Tokmakov17b,Tokmakov17a}, and object saliency~\cite{Li18g} using motion and appearance related cues.

Motion-based methods~\cite{Lee11,Papazoglou13,Factor14,Keuper15,Wang15d,Haller17,Koh17b,Stretcu15,Yang19c} define the foreground object based on the region that moves differently from the rest of the scene, and they integrate this supervision through optical flow images and temporal consistency.~\cite{Koh17b} combines the flow information with the recurrence property of the primary object in an image sequence and identifies the matching segment tracks across frames by extracting ultrametric contour maps. Similar to our approach,~\cite{Stretcu15} assumes that the foreground is harder to model than the background and while modeling the background by a low-dimensional linear basis, the image parts this model fails to explain are identified as the salient object. In contrast to~\cite{Stretcu15}, our method relies on the predictability of image patches from their spatial neighborhood using deep neural networks, can handle complex background motion and does not require videos. Built upon~\cite{Stretcu15},~\cite{Croitoru19} trains an ensemble of networks, which comes at the cost of requiring significant amounts of additional data. In~\cite{Katircioglu20}, an inpainting network is trained to identify the regions that are harder to reconstruct from the surrounding image patches and encodes and decodes the content of this region to learn the scene decomposition. \cite{Yang19c} employs a similar inpainting network but on flow fields obtained by~\cite{Sun18a} and aims to generate the mask of a moving object in the region where the inpainting network yields poor reconstruction. Methods based on deep optical flow are not strictly self-supervised and can yield degenerate solutions when applied to still images with no or little movement. Recently,  temporal information at different granularities has also been used via forward-backward patch tracking~\cite{Lu20}. Note that these methods can only operate on video streams and exploit a strong temporal dependency, which our model doesn't.

Recent self-supervised methods that operate on single RGB images employ generative models to detect the regions that can be exposed to certain transformations without changing the realism of the image~\cite{Bielski19,Chen19a,Arandjelovic19,Benny20}. However, these methods can easily be deceived by other background objects whose random displacement or texture change can still yield realistic images. In contrast to all of these techniques, our approach works with single images acquired using a moving camera and with an arbitrary background.

\parag{Multi-View Self-Supervised Approaches.}
Other relevant approaches include the generative unsupervised multi-person detection and tracking methods proposed in~\cite{Fleuret08a, Baque17b}. The former localizes and matches persons across several cameras with overlapping fields of view using a grid of candidate positions on the ground plane. The latter uses a joint CNN-CRF architecture and Mean-Field inference to produce a Probabilistic Occupancy Map (POM) as in~\cite{Fleuret08a} but leverages discriminative features extracted by a CNN. Both require background subtraction images as input and can therefore only work with static cameras. Furthermore, they exploit multiple views at inference time, whereas we aim to perform monocular person segmentation.

\parag{Multi-View Self-Supervised Training for Single View Inference.}
Our work is closely related to~\cite{Rhodin18b,Rhodin19a} in that we do not use any segmentation annotation to learn the foreground region. In~\cite{Rhodin18b,Rhodin19a}, novel view synthesis is used in conjunction with multi-view synchronized videos of human motions captured by calibrated cameras to learn a geometry-aware embedding. In contrast to our approach, it requires a known background to decompose the scene into foreground and background regions. Hence, it cannot handle scenes filmed by moving cameras. Here, we introduce a method that works with a changing background. To this end, we do not rely on novel view synthesis but instead exploit multi-view consistency by relating the 2D detections of the multiple views to a common 3D capture volume.


\section{Method}

Our goal is to develop a self-supervised algorithm that generates a bounding box and the corresponding segmentation mask from a single image. However, whereas earlier methods use videos from a single camera for training purposes, we want to demonstrate that using calibrated and synchronized cameras for training purposes increases performance. Therefore, let us assume that we have  videos acquired by $C > 1$ calibrated and synchronized projective cameras. For each $c$ between 1 and $C$, camera $c$ captures image $\mI_c$ and its behavior is modeled by a $3 \times 4$ projection matrix $\mP_c$.

\subsection{Multi-View Self-Supervised Training}
\label{multi-view training}

Let us now turn to the task of exploiting such multi-view data to train our detection and segmentation network. Because we ultimately aim to perform single-view 2D detection and segmentation, our approach produces bounding boxes and segmentation masks for each individual view. Nevertheless, we exploit multi-view geometry to better constrain the training process and enforce consistency across the views. Furthermore, we do this without requiring additional loss terms that would make the process more complex and force us to carefully weigh these additional terms against the original ones. To this end, our training algorithm goes through the following steps
\begin{enumerate}

\vspace{-1mm}
\item We use a network $\cF$ to compute a probability map for 2D bounding boxes over an image grid for each view $c$. These probability maps are used to vote in a 3D grid for potential 3D locations of these bounding boxes.

\vspace{-1mm}
\item We sample individual 3D voxels in that 3D grid according to the resulting probability density. This corresponds to one 2D bounding box for each view.

\vspace{-1mm}
\item We compute the 3D center and object height that best agree with these 2D bounding boxes in a least-square sense.

\vspace{-1mm}
\item We project the resulting 3D center and height in each view to define new 2D bounding boxes, keeping the original width of the sampled boxes.

\vspace{-1mm}
\item These boxes are then used to evaluate the loss function associated to $\cF$ in each image.

\end{enumerate}
Multi-view consistency is achieved both by sampling the 3D proposal grid and adjusting the 2D bounding boxes. Hence, we do not require additional losses to enforce consistency. This is a central element of our approach because, as  observed in~\cite{Rhodin18a}, such loss terms tend to favor degenerate solutions that are consistent but wrong. This is something our ablation study confirms. In the remainder of this section, we describe these steps in more detail. 


\subsubsection{Bounding Boxes in Individual Views}
\label{sec:box2D}


\begin{figure}
	\begin{center}
		\includegraphics[width=\linewidth]{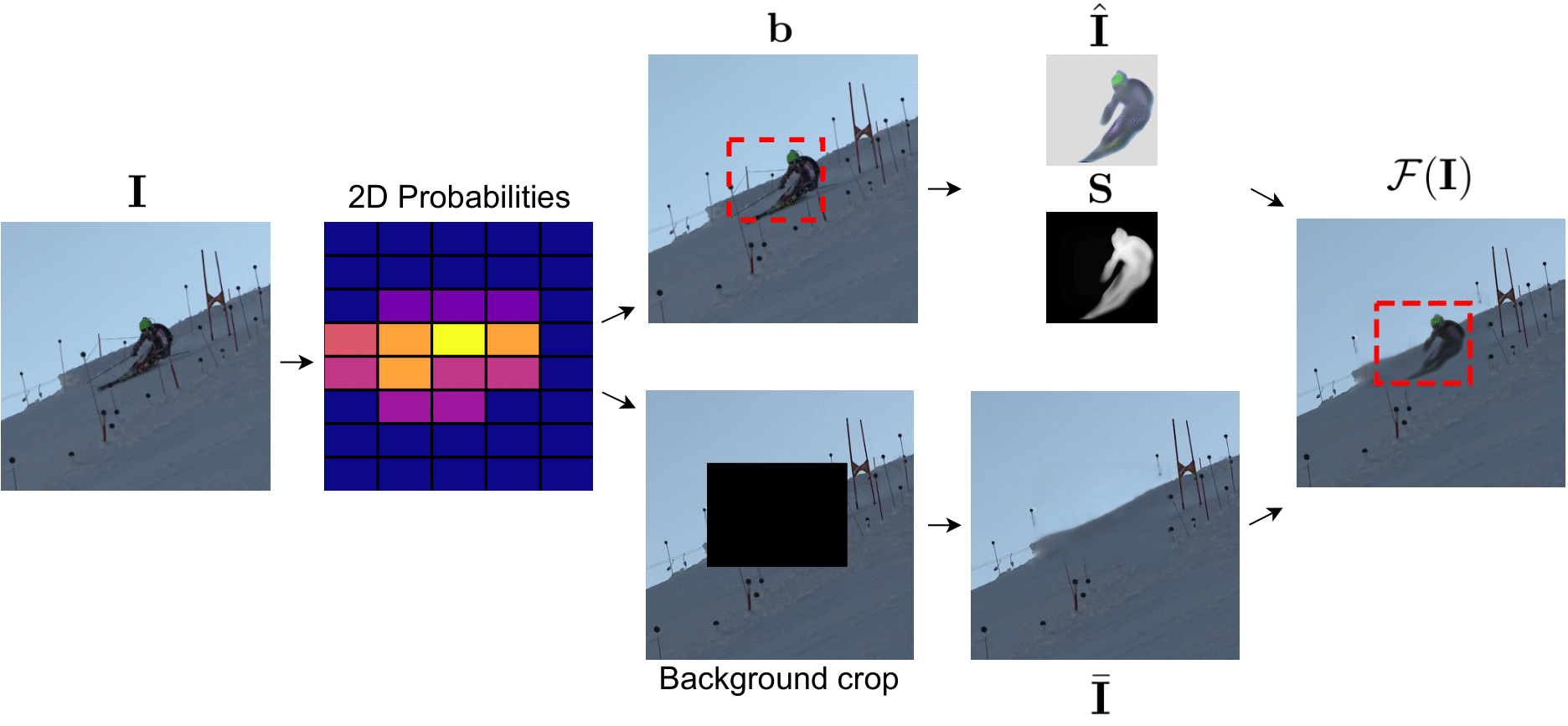}
	\end{center}
\vspace{-4mm}
	\caption{\label{fig:summary2d}
		{\bf Single-view self-supervised segmentation.} This figure summarizes our starting point, the single view approach. It predicts 2D occupancy probabilities, an associated bounding box, and a foreground mask within this window. It is trained to reconstruct the input image by pasting the foreground region underneath the mask on a background image obtained by inpainting the predicted bounding box.
	}
\vspace{-4mm}
\end{figure}

Let us consider the network $\cF$ of~\cite{Katircioglu20}, which we use as the backbone of our approach. It takes an image $\mI \in \mathbb{R}^{W \times H \times 3}$ as input and resynthesizes it. In the process, it produces a probability map over a grid, encoding for each cell $i$ the probability $p_{i}$ that a bounding box ${\vb}_{i}$ at this location contains a person. 
As depicted by Fig.~\ref{fig:summary2d}, resynthesis is achieved by sampling a candidate bounding box, cropping the corresponding image patch, and, in parallel, predicting a foreground image $\hat{\mI} \in \R^{128 \times 128 \times 3}$ and a segmentation mask $\mS \in \R^{128 \times 128}$ from the crop, while inpainting the cropped region to generate a background image $\bar{\mI}  \in \R^{W \times H \times 3}$. We then re-compose the foreground crop and the background image according to the segmentation mask. Formally, this can be written as 
\begin{align}
\cF(\mI) =&  \cT^{-1}(\hat{\mI} \circ \mS) + \bar{\mI} \circ (1 - \cT^{-1}(\mS)), \label{eq:merge_eqn}
\end{align}
where $\cT$ is the spatial transformer corresponding to the selected bounding box, and $\circ$ is the element-wise multiplication. This allows one to train $\cF$ in a self-supervised fashion, by comparing the reconstructed image to the input one.


\subsubsection{Consistent Sampling using a 3D Proposal Grid}
\label{sec:grid}

To link 2D detections across views, we construct a 3D proposal grid with $V$ voxels centered at the point nearest to the optical axes of all cameras in the 3D world coordinate system, as shown in Fig.~\ref{fig:Grid3D}.  For each voxel $j$ of that grid, we compute its center $\vv_j \in \mathbb{R}^{3}$, together with a probability of occupancy  $q_j$, discussed below.

Since we know the camera matrix $\mP_c$ for each image $\mI_c$, we can project the center $\vv_j$ of each 3D voxel into it. The projected center will fall into image grid cell $i^c(j)$ to which $\cF$ has associated a probability $p^c_{i^c(j)}$, as discussed at the beginning of Section~\ref{sec:box2D}. We repeat this operation over all images and all voxels and sum the resulting log probabilities for each voxel.  We then normalize the resulting probability density over the 3D grid so that it integrates to one. Formally, this can be written as
\begin{equation}
q_{j} = \frac{1}{Z}\exp\left(\sum_{c} \log(p^c_{i^{c}(j)})\right) \; ,
\label{eq:probabilityReconstruction}
\end{equation}
where $Z$ is a normalization constant easily computed on a discrete grid of finite dimensions. 

To train our network in a self-supervised fashion, we then sample one voxel location $j$ according to the distribution in Eq.~\ref{eq:probabilityReconstruction}. The sampled voxel then corresponds to one bounding box candidate in each view, inherently encouraging consistency across the views as illustrated in Fig.~\ref{fig:multiview}(a). This consistency, however, is only a partial one because each view still predicts the precise location and dimensions of its own bounding box. Hence, the final bounding boxes may still disagree. To prevent this, we explicitly enforce geometric consistency as discussed below.

\subsubsection{Enforcing Geometric Bounding Box Consistency}
\label{sec:consistency}

\begin{figure}[t]
	\centering
	\begin{tabular}{c}
		\includegraphics[width=0.35\textwidth]{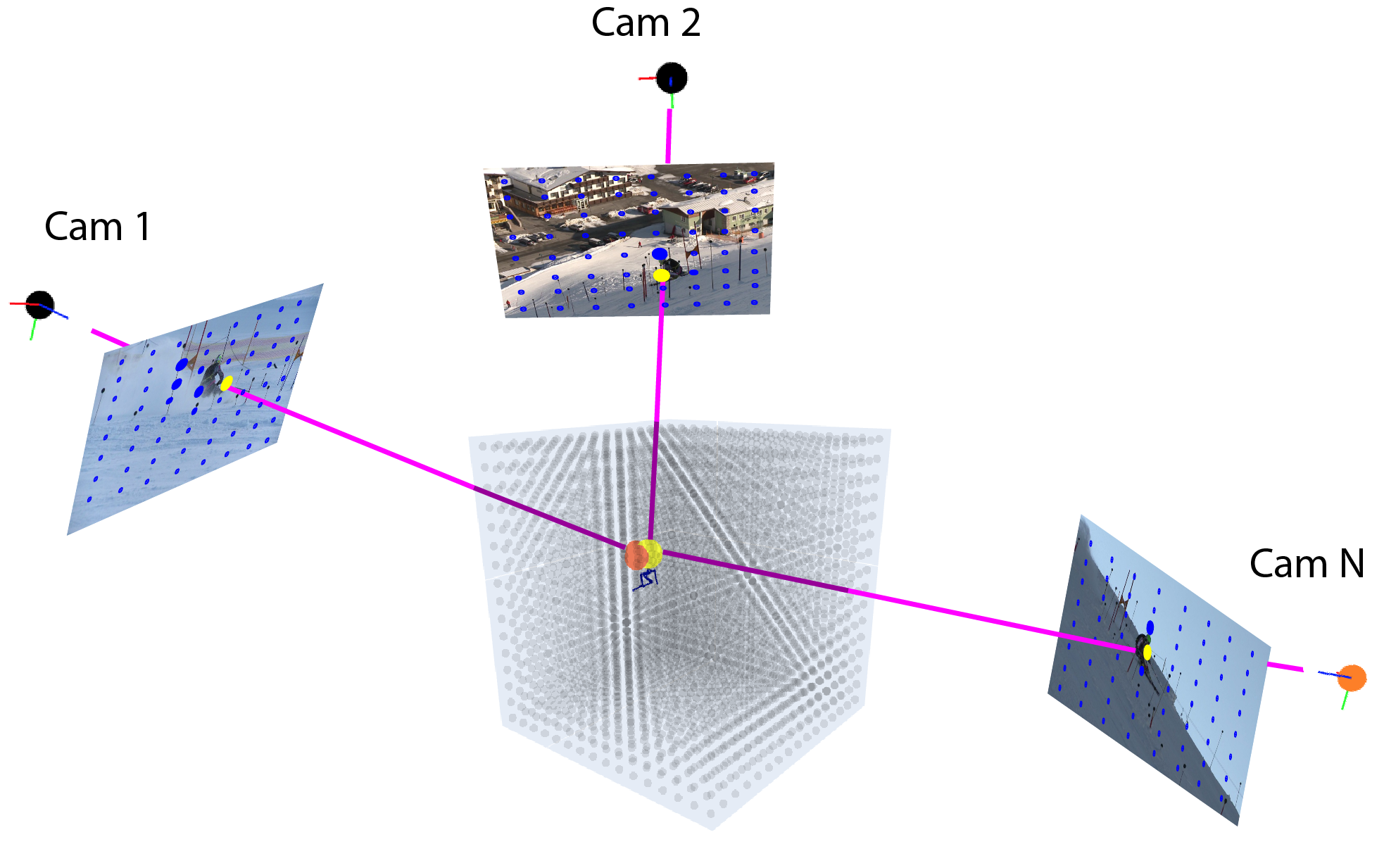}\\
		(a) Multi-view voting. \\
		\includegraphics[width=0.35\textwidth]{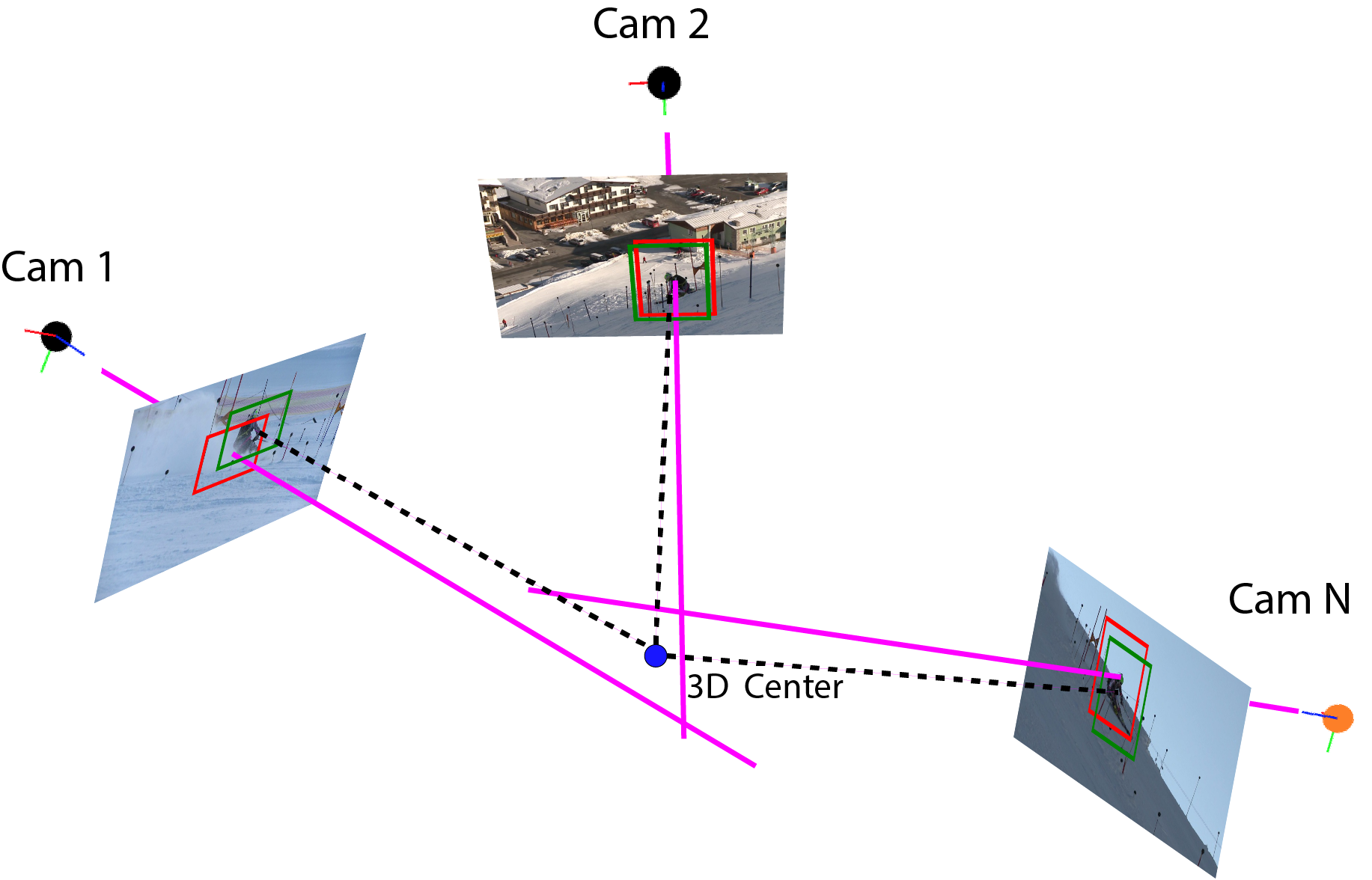}\\
		(b) Making the bounding box centers consistent. \\
			\includegraphics[width=0.35\textwidth]{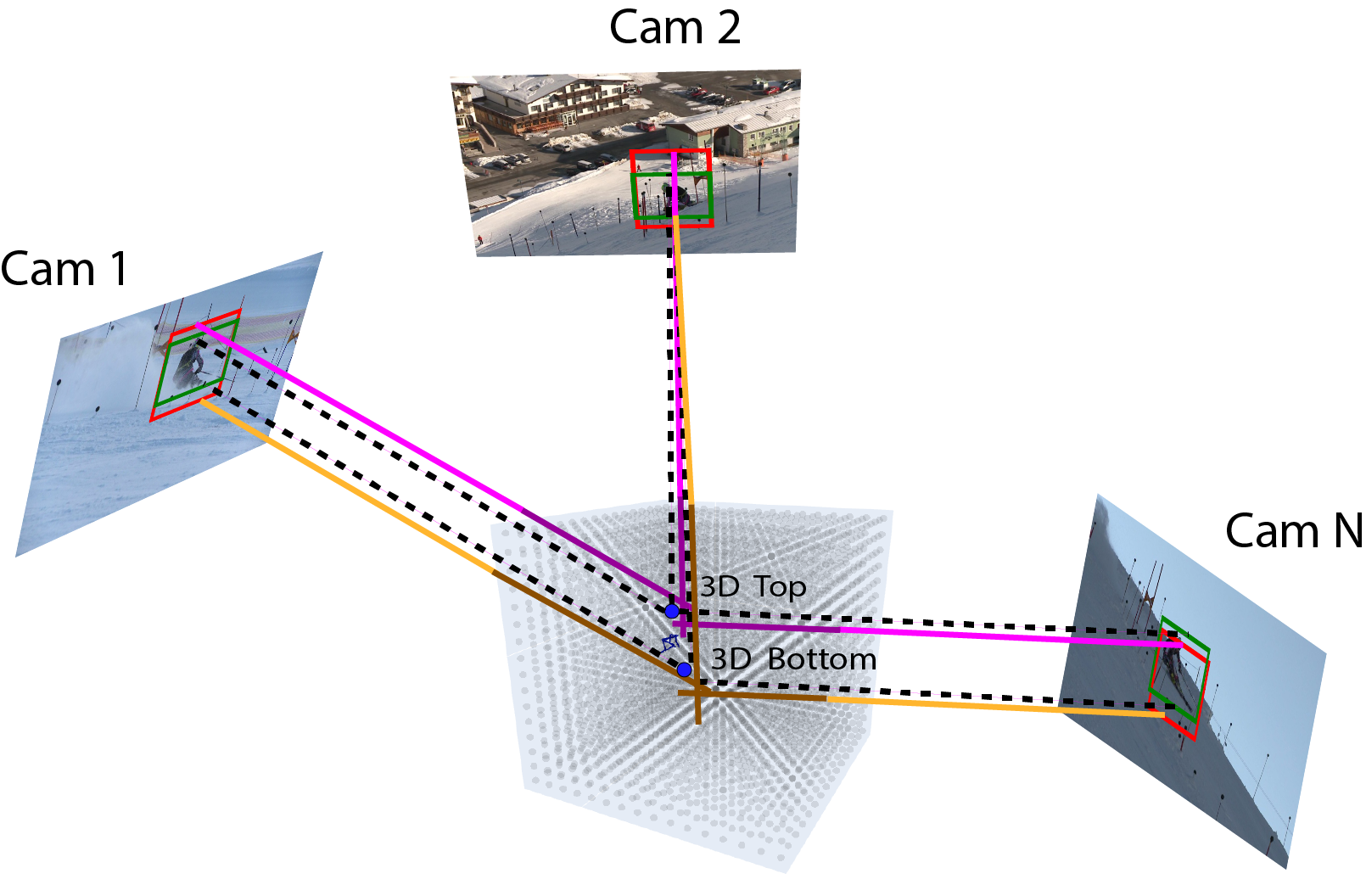}\\
		(c) Making the bounding box heights consistent. \\
	\end{tabular}
	\caption{\textbf{Finding bounding boxes that are view consistent.} (a) The blue dots overlaid on each view represent the initial 2D probabilities and vote in the 3D grid along their respective lines of sight. As a result, the yellow 3D voxel becomes very likely to be sampled. (b) The red bounding box drawn in each view is the initial prediction and the purple line of sight is going through the bounding box center. The 3D center is the point closest to all these lines and its re-projection in the images becomes the center of the new bounding boxes, shown in green. (c) The red bounding boxes represent the initial prediction and the purple and orange lines indicate the line of sight going through the bounding box top and bottom points. The 3D top and bottom locations are taken to be the point closest to purple and orange lines respectively. Their re-projection in the images become the top and bottom middle points of the new bounding boxes, shown in green.}%
	\label{fig:multiview}
	\vspace{-6mm}
\end{figure}

To enforce geometric consistency between the bounding boxes from different views, we want to ensure that their 2D centers all match the same point in 3D and that their 2D heights correspond to the same 3D size. In other words, we want to modify the bounding box locations so that the new ones have consistent 2D centers and heights and we want to achieve this with as little displacement as possible. Since the cameras are often set in a rough circle pointing at the subject, enforcing height consistency makes sense because the camera up directions are aligned. Only when the camera angle varies, as in drone footage taken from arbitrary angles, should the height constraint be replaced. We do not constrain the bounding-box width because the left-right direction of cameras is not aligned unless the cameras are parallel. This makes the width view dependent, as in Fig.~\ref{fig:teaser} where the skier's projection is wider in some views.

In essence, we seek to {\it project} the bounding boxes to new ones that satisfy the center and height constraints and that will be used by the network to evaluate its objective function during its forward pass. It is therefore essential that this projection be differentiable such that the backward pass can be carried out during training. 

\parag{Adjusting  Bounding Box Centers.} 

As shown in Fig.~\ref{fig:multiview}(b), we use the lines of sights defined by the 2D centers of the bounding boxes, find the 3D point closest to all of them, and use its re-projection into the images as the modified center for the bounding boxes. For each view $c$, the line of sight $\vl_c$ in image $\mI_c$ can be expressed 
as 
\begin{equation}
\vl_c = \mM_c^{-1} \left[u_c , v_c ,1 \right]^T \; , 
\label{eq:center_backproj}
\end{equation}
where $\mM_c$ is the $3 \times 3$ matrix formed by the first 3 columns of $\mP_c$ and $u_c,v_c$ are the 2D pixel coordinates of the bounding box center in $\mI_c$. Hence, finding the point closest to all the $\vl_c$ amounts to solving a least-squares problem, which can itself be achieved by solving a linear system of equations and is therefore differentiable. In practice, we use a differentiable least-squares implementation for this purpose and provide its details in the supplementary material.

\parag{Adjusting Bounding Box Heights.} 

As shown in Fig.~\ref{fig:multiview}(c), we similarly use the midpoints of the top and bottom parts of the bounding boxes in each view to predict two new intersection points, one for the top and one for the bottom of the bounding box in 3D. We then take the distance between the re-projections of these points into the image to be the new height of the bounding boxes.  As before, this is a differentiable operation.



\subsubsection{Training}
\label{sec:training}

Because our 2D bounding boxes are made to be consistent, we can train our network by minimizing the same loss as in the single view approach of~\cite{Katircioglu20}, except for the fact that we jointly compute it over several images, and do not require to introduce an additional loss to enforce consistency. 

More specifically, we minimize the weighted sum of two loss functions $\objBG(\mI_1, \ldots, \mI_C)$ and $\objFG (\mI_1, \ldots, \mI_C)$. $\objBG$ accounts for the fact that a region containing a moving foreground object is unlikely to be well re-synthesized by the inpainter and is critical to train the network to place the bounding box at the right location in each image. $\objFG$ gauges how well $\cF$ resynthesizes the complete original images and is minimized when the segmentation mask fits the salient object as well as possible within the sampled bounding box. In practice, they are taken to be
\begin{align}
\objBG(\mI_1, \ldots, \mI_C)  & = - \sum_{c=1}^{C} r_j \frac{\|\bar{\mI}_c - \mI_c \|^2}{area(\vb^c_{i^{c}(j)})} \; ,  \label{eq:background objective} \\
\objFG (\mI_1, \ldots, \mI_C) & = \sum_{c=1}^{C} r_j \| \cF(\mI_c) - \mI_c \|^2 \; , \label{eq:foreground objective}
\end{align}
where $ area(\vb^c_{i^{c}(j)}) \in \mathbb{N}_0$  is the area of the bounding box obtained by sampling voxel $j$ and enforcing geometric consistency. As in~\cite{Katircioglu20}, the sampled voxel is obtained by importance sampling, and $r_j$ is the ratio of the probability $q_j$, from Eq.~\ref{eq:probabilityReconstruction}, by its importance sampling probability. In addition to these loss terms, as~\cite{Katircioglu20}, we use an $L_1$ prior on $\mS$ to favor a crisp segmentation, and compute Eq.~\ref{eq:foreground objective} not only on pixel color but also on learned features. Additional details on the sampling, hyper-parameters, training and network architectures are provided in the supplementary material.



\subsection{Single-View Inference}

Once trained using multiple views, our model can detect and segment the salient object from single RGB images at inference time without any further changes. We run our network on the image and simply choose the 2D grid cell with the highest occupancy probability. Its bounding box parameter estimations are fed into the spatial transformer $\cT$ to crop the region of interest, which is encoded into the corresponding segmentation mask and foreground, and decoded into the reconstructed image as illustrated in Fig.~\ref{fig:summary2d}.

\comment{

\subsection{Network Architecture}

The architecture of $\cF$ must allow for self-supervision. To this end, we therefore base ours on the two-stream one of~\cite{Katircioglu20}. It is depicted by Fig.~\ref{fig:summary2d}, and operates as follows.  

First, a YOLO-like network component~\cite{Redmon16} predicts a set of $N$ candidate object locations $\{{\vb}_{i}\}^N_{{i}=1}$ with corresponding probabilities $\{p_{i}\}^N_{i=1}$. One bounding box is then sampled based on the $\{p_{i}\}$ probabilities. This sampling is not naturally differentiable but differentiation can nevertheless be achieved using  a Monte-Carlo based strategy. The selected bounding box is fed to a first branch, shown at the top of Fig.~\ref{fig:summary2d}, that encodes and subsequently decodes the image within the bounding box into a foreground image $\hat{\mI} \in \R^{128 \times 128 \times 3}$ and a segmentation mask $\mS$ using a bottleneck autoencoder $\cS$. \MS{Can you include the notation in Fig.~\ref{fig:summary2d}?}The selected bounding box is also fed to a second branch that crops it from the original image and uses an inpainter to fill the resulting gap and produce a background image $\bar{\mI} \in \R^{W \times H \times 3}$, as shown at the bottom of Fig.~\ref{fig:summary2d}. $\hat{\mI}$ and $\bar{\mI}$ are then combined \ms{according to the predicted segmentation mask $\mS$, by undoing the crop of $\hat{\mI}$ using an inverse transformer,}
to produce a final image $\cF(\mI)$ that should closely resemble the original one $\mI$. Formally, this can be written as 
\begin{align}
\cF(\mI) =&  \cT^{-1}(\hat{\mI} \circ \mS) + \bar{\mI} \circ (\cT^{-1}(1 - \mS)), \label{eq:merge_eqn}
\end{align}
where $(\hat{\mI},\mS) =  \cS(\cT(\mI))$, $\cT$ is the spatial transformer corresponding to the selected bounding box, and $\circ$ is the element-wise multiplication. 

To train the network in an unsupervised fashion, the weighted sum of two loss functions $\objBG(\mI)$ and $\objFG (\mI)$ is minimized. $\objBG$ accounts for the fact that a region containing a moving foreground object is unlikely to be well re-synthesized by the inpainter and is critical to train the network to place the bounding box at the right location. $\objFG$ gauges how well $\cF$ resynthesizes the whole original image and is minimized when the segmentation mask fits the salient object as well as possible within the sampled bounding box. In practice, they are taken to be
\begin{align}
\objBG(\mI)  & = -\frac{\|\bar{\mI} - \mI\|^2}{area(\vb_{i})} \; ,  \label{eq:background objective} \\
\objFG (\mI) & = \| \cF(\mI) - \mI \|^2 \; , \label{eq:foreground objective}
\end{align}
where $area(\vb_{i}) \in \mathbb{Z}$ \MS{Why $\mathbb{Z}$? Is the area always an integer? Even so, it is always non-negative, so it should be $\mathbb{N}_0$, no?} is the area of the selected bounding box. 

\PF{Probably should mention how $\objBG$ and $\objFG$ are used in conjunction with the consistency loss functions introduced above.}

\PF{End of edits}

\subsection{Single-View Self-Supervised Training}

\subsection{Multiple-View Self-Supervised Training}
\label{sec:multiview_consistency}

Let us now assume that we have multiple videos acquired by calibrated and synchronized cameras. We could train the network $\cF$ on each view independently but that would fail to exploit the fact that detections should be geometrically consistent across views. A simple way to remedy this would be to train the networks jointly and add loss terms that favor consistency across views. We have tried this and found that such loss terms tend to favor degenerate solutions that are consistent but wrong, as already observed in~\cite{Rhodin18a}. Instead, we use the 2D box proposals that $\cF$ yields from individual images to vote into a 3D proposal grid, as shown in Fig.~\ref{fig:3Dsetup}. We then exploit multi-view geometry to enforce consistency of these backprojections. We use a Monte Carlo sampling strategy that makes the 3D grid voting differentiable and allows us to train the resulting network. 


\paragraph{Camera Models.}  

Let us consider $C$ perspective cameras and, for the sake of compactness, let us express all 2D and 3D points in homogeneous coordinates. In other words, we represent 2D points as vectors in 
$\mathbb{R}^3$ and 3D points as vectors in $\mathbb{R}^4$. 

Camera $c$ can be modeled by a $3 \times 4$ projection matrix $\mathbf{P}^c = [ \mathbf{M}^c | \mathbf{m}^c ]$, where $ \mathbf{M}^c$ is a $3 \times 3$ matrix and $ \mathbf{m}^c$ a 3-vector. Given a 3D point $\mathbf{X} \in \mathbb{R}^{4}$, its projection in image $c$ is $\mathbf{x} = \textbf{P}^{c} \mathbf{X} \in \mathbb{R}^{3}$. \MS{Should we really use an = sign here?} Conversely, given a 2D point $\mathbf{x} \in \mathbb{R}^{3}$, the line of sight, that is, the line formed by all the points that can project there, is the set of points 
\begin{align}
\mathbf{X}    &= \mathbf{O}^{c} + \lambda \left[ \begin{array}{c}(\mathbf{M}^c)^{-1} \mathbf{x} \\ 0 \end{array} \right] , \quad \forall \lambda \in \mathbb{R}  \; , \label{eq:lineofsight}  
\end{align}
\MS{Is it correct to have the rotation taken into account in the inverse, i.e., not just the intrinsic parameters?}
where $\mathbf{O}^{c}$ is the optical center of the camera. 


\paragraph{3D Proposal Grid.}  

We construct a 3D grid of proposals around a center location, taken to be point nearest to the optical axes of all cameras in the 3D world coordinate system. We parameterize the grid with $X \times Y \times Z$ voxels, each voxel $j$ having an associated center position $\mathbf{V}_j \in \mathbb{R}^{4}$ and occupancy probability $q_j$. 

\PF{That would be the place to explain how bounding boxes are backprojected into this proposal grid, given the camera models described above.}

\paragraph{Geometric Relation of the 3D and 2D Proposal Grids.} 

Given a perspective pinhole camera model with 
\begin{equation}
\begin{aligned}
\begin{bmatrix}
u \\
v \\
1 \\
\end{bmatrix}
= \textbf{P}^{c} \mathbf{V}_j 
\end{aligned}
\label{eq:voxel_projection}
\end{equation}
\MS{Again, should we really use a strict equality?}
where $u$ and $v$ are the pixel coordinates that the center of voxel $j$ corresponds to. The 2D grid cell index $i^{c}(j)$ corresponding to voxel $j$ can now be derived as
\begin{equation}
i^{c}(j) = \lfloor (u \, N_{col}) / W\rfloor + N_{col} \lfloor(v \, N_{row}) / H \rfloor\;,
\label{eq:cell projection}
\end{equation}
where $N_{col}$ and $N_{row}$ are the 2D grid sizes. 

Conversely, the 3D position can be triangulated from multiple 2D estimates. 
We backproject the projective coordinates of the 2D image location $(u^c,v^c)$ of view $c$ using Eq.~\ref{eq:lineofsight}.


\paragraph{Probabilistic Relation of the 3D and 2D Proposal Grids.}
Multiple 3D voxels project onto the same 2D cell. In this probabilistic framework, the probability of the person being in the 2D cell $i^c$ is the 3D probability marginalized over the depth direction from each camera view. It can be written as
\begin{equation}
{p^{c}_{i}} = \sum_{j\text{ with } i^{c}(j) = i } q_j,
\end{equation}
where the sum is over all 3D cells $j$ that project on 2D cell $i$. Conversely, the 3D probabilities can be reconstructed from the marginals of multiple views as
\begin{equation}
q_{j} = \frac{1}{Z}\prod_{c} p^c_i, \text{ with } i={i^{c}(j)},
\label{eq:probability reconstruction naive}
\end{equation}
with $Z$ a normalization constant that is easily computed on the discrete grid of finite dimensions. The reconstruction of the joint distribution from its marginals is in general an ill-posed problem. In Eq.~\ref{eq:probability reconstruction naive}, we exploit the fact that we have a single person in the scene such that no occlusion can occur and the distribution can be assumed to have a single mode. In practice, we use the numerically more stable variant
\begin{equation}
q_{j} = \frac{1}{Z}\exp\left(\sum_{c} \log(p^c_i)\right), \text{ with } i={i^{c}(j)}.
\label{eq:probability reconstruction_old}
\end{equation}
%
%


%


\paragraph{Multi-view Consistent Bounding Box.}

We introduce three different ways to predict the precise location of the bounding box in all camera views. The baselines are illustrated in Fig. \ref{fig:multi_view_consistency_baselines}.\\
\texttt{1.Proposals3D}: In this version of our method, only the proposals are made consistent in 3D across the views. As stated before, we use the detection network $\cD$ to predict the bounding box vector $\vb^c_{i} = [g_{i,x}+\delta x, g_{i,y}+\delta y, s_x, s_y]$ and 2D probabilities $p^{c}_i$, or rather $\log(p^{c}_i)$, individually for every camera $c$.  The 2D bounding box is parameterized as the box width and height, $s_x, s_y \in [0,1]$, and the offsets, $\delta x, \delta y \in [-1, 1]$, from the cell pixel location, $g_{i,x}, g_{i,y}$. Our contribution is the linking of the individual predictions on the 3D voxel grid. First, a consensus is found in 3D by reconstructing the joint probability distribution from the 2D marginals using Eq.~\ref{eq:probability reconstruction}.  To facilitate optimization of this distribution, we sample from it using importance sampling, which yields a 3D cell $j$ and corresponding 2D cells in each camera via the projection formula. This sampling in 3D forces the network to predict offsets and 2D probabilities that agree, thereby favoring positions on the foreground since background objects erroneously selected in one view are unlikely to agree with salient objects in all other views. Below, we explain how we further make fine-grained offset prediction and bounding box size regression consistent across the views.\\
\texttt{2.Center3D-Proj}: The same vector $\vb^c_{i}$ is predicted per view, but now the bounding box centers in each view are triangulated to estimate a single 3D point from all offsets. The center location of the sampled proposal in camera $c$ can be written as
\begin{equation}
\begin{aligned}
{u}^{c} &= W * (\delta x + 1) /2 + g_{x} \\
{v}^{c} &= H * (\delta y + 1) /2 + g_{y} 
\label{eq:bbox center}
\end{aligned}
\end{equation}
where ${u}^{c} \in [0, W]$ and ${v}^{c} \in [0, H]$ are in pixel coordinates. The 3D location of the proposal center $\vu^c$ in world coordinates is calculated using Eq.~\ref{eq:lineofsight}. To reach a multi-view consensus on the center of a 3D bounding box, namely $\textbf{\=u} \in {\mathbb{R}}^{3 \times 1}$, we take into account the lines emerging from camera optical center $\mathbf{O}^{c} \in {\mathbb{R}}^{3 \times 1}$ and passing through $\vu^c$ for each camera. Since these lines are unlikely to agree perfectly, we compute the point that is closest to all these 3D lines in the least squares sense. We make this step differentiable by building a system of linear equations and using a differentiable least squares solver. The unit direction vector for each line is
\begin{equation}
\begin{aligned}
\textbf{n}^c &=  (\vu^c - \mathbf{O}^{c} ) / \norm{\vu^c - \mathbf{O}^{c}},
\label{eq:unit_direction}
\end{aligned}
\end{equation}
where $\textbf{n}^c \in  {\mathbb{R}}^{3 \times 1}$. To find the nearest point $\textbf{\=u}$ to a set of lines, we calculate the point with minimum distance to them. Given that each line is defined by its origin $\mathbf{O}^{c}$ and the unit direction vector $\textbf{n}^c$, the squared perpendicular distance from the point $\textbf{\=u}$ to one of these lines is given by
\begin{equation}
\begin{aligned}
\textbf{d}^c &=  (\mathbf{O}^{c}  - \textbf{\=u})^{T} (\bf I - \textbf{n}^c(\textbf{n}^c)^T)  (\mathbf{O}^{c}  - \textbf{\=u})
\label{eq:distance}
\end{aligned}
\end{equation}
where the matrix $ (\bf I - \textbf{n}^c(\textbf{n}^c)^T) $ serves as the projector of the line vectors into the space orthogonal to $\textbf{n}^c$. By minimizing the sum of squared distances, we can obtain the nearest point in the least squares sense. To this end, we solve a linear system of equations
\begin{equation}
\begin{aligned}
\bf A \textbf{\=u} &= \bf m ,\\
\bf A &= \sum_{c=1}^{N} (\bf I - \textbf{n}^c(\textbf{n}^c)^T), \\
\bf m &= \sum_{c=1}^{N} (\bf I - \textbf{n}^c(\textbf{n}^c)^T) \mathbf{O}^{c},
\label{eq:lstsq}
\end{aligned}
\end{equation}
with $\bf A \in {\mathbb{R}}^{3 \times 3}$ and $\bf m \in  {\mathbb{R}}^{3 \times 1}$.  The optimum is achieved at the least squares solution. Therefore, $\textbf{\=u} = lstsq(\bf A, \bf m)$ and we use the differentiable PyTorch $\textit{lstsq}$ function to solve it. 

The re-projection of $\textbf{\=u}$ onto each view is used as the new crop location in the forward pass. Thus, 
\begin{equation}
\begin{aligned}
\begin{bmatrix}
{\bar{u}}^c \\
{\bar{v}}^c \\
1
\end{bmatrix}
= \textbf{P}^{c} \textbf{\=u}
\end{aligned}
\label{eq:new_center_2D}
\end{equation}
where $ \textbf{\=u}$ is the new center and $ [{\bar{u}}^c, {\bar{v}}^c]$ are the updated 2D bounding box centers in each view. In addition to the 3D proposals, this forces the network to predict a precise bounding box that is consistent across all views, thereby avoiding trivial solutions in any single view. We provide more details in the supplementary material. \\
\texttt{3.BBox3D-Proj}: We predict an additional pair of offsets to define the top and bottom location of the bounding box as 
The top and bottom locations, $[{u}^{t,c}  , {v}^{t,c} ]$ and $[{u}^{b,c}, {v}^{b,c}]$ respectively, of the bounding box in camera view $c$ are computed as
\begin{equation}
\begin{aligned}
{u}^{t,c} &= W * (\delta x + 1) /2 + g_{x} \\
{v}^{t,c} &= H * (\delta y + 1) /2 + g_{y} - (H * s_x)/2 \\
{u}^{b,c} &= W * (\delta x + 1) /2 + g_{x} \\
{v}^{b,c} &= H * (\delta y + 1) /2 + g_{y} + (H * s_x)/2
\label{eq:bbox top_bottom}
\end{aligned}
\end{equation}
The 3D locations of the top and bottom points of the bounding box, $\textbf{u}^{t,c}$ and $\textbf{u}^{b,c}$ in camera view $c$ can be computed using Eq.~\ref{eq:lineofsight}. We apply the same least squares triangulation and re-projection as the previous baseline, and take their difference along the vertical direction to compute the bounding box height. This third version ensures that both height and center position are consistent across the views. Although a similar strategy could be applied for the bounding box width, we found view-dependent width estimates, details of which are included in the ablation study, to be more accurate in practice.

\paragraph{Training}
Using this 3D grid, we define an object function that is the sum of monocular objectives similar to the one of Eq.~\ref{eq:foreground objective}. Specifically, we write our 
objective as
\begin{equation}
\begin{aligned}
\objFG(\mI) = \sum_{c} \|\cF(\mI) - \mI\|^2, \text{ with} \ {j} \sim q_j \;, 
\end{aligned}
\label{eq:multi-view objective}
\end{equation}
where the probability is now maximized over the 3D grid, with probability $q_j$ for voxel $j$ found via a consensus-voting scheme, and the bounding boxes $\vb^c_{j}$ are re-projected after triangulating the initial view-dependent estimates and enforcing multi-view consistency. 

\subsection{Single-View Inference}
Once trained on multiple views, our model can detect and segment the salient object from single RGB images at inference time without any change. Instead of sampling from the 3D voxel proposals to select a 2D grid cell to yield a bounding box, we directly choose the 2D grid cell with the highest occupancy probability. Its bounding box parameter estimations are fed into the spatial transformer $\cT$ to crop the region of interest, which is encoded and decoded into the reconstructed image via $\cS$. 

}

\section{Experiments}
\label{sec:eval}

Unlike that of~\cite{Rhodin19a},  our self-supervised approach is designed to work using multiple-cameras that can move. In this section, we show that it does, yet outperforms~\cite{Rhodin19a} even when the background is static. Furthermore, we show that using multiple cameras for training purposes delivers the hoped-for performance boost over state-of-the-art monocular approaches~\cite{Koh17b,Yang19c,Croitoru19,Lu20,Katircioglu20}.

\subsection{Images and Metrics}

We first describe the image datasets we work with and then the metrics we use for comparison purposes. 

\parag{Images acquired using moving cameras.}

The \ski{} dataset of~\cite{Rhodin18a} features six skiers on a slalom course. We use the official training/validation/test sets that split the 12 videos of six skiers as four/one/one, with, respectively, $7800$, $1818$ and $1908$ frames. The pan-tilt-zoom cameras constantly adjust to follow the skier. Nothing remains static, the background changes quickly, and there are additional people standing in the background. The cameras were calibrated using static scene markers {\it without} any markers or keypoints on the skier's body. We use the full image as input and evaluate detection accuracy using the available 2D pose annotations and segmentation accuracy of the $300$ labeled frames in the test sequences. To pick the hyperparameters, we use $36$ labeled validation frames ($3$ frames each from six cameras and two sequences). Due to the large distance between cameras and subject, the 3D proposal grid has $16^3$ voxels with cuboid side length of 8 meters.

To demonstrate the applicability of our method to scenes without an initial camera calibration,  we use the \handheld{} dataset~\cite{Katircioglu20} captured by three hand-held cameras that translate and rotate in an unscripted fashion. It comprises three training, one validation, and one test sequences. They all feature one person  performing actions mimicking the human{} motions in an outdoor environment with a changing background. We used OpenSFM \footnote{https://www.opensfm.org/} to calibrate 4200 frames from the training set using and tested on the same images as~\cite{Katircioglu20}. The 3D proposal grid has $16^3$ voxels with cuboid side length of 12 meters.


 
\parag{Images acquired using Static Cameras.}

To compare against algorithms requiring a static background, we evaluate our approach in the more controlled environment of the \human{} dataset~\cite{Ionescu14a}. It was acquired using four static cameras and comprises 3.6 million frames and 15 motion classes. It features 5 subjects for training and 2 for validation, seen from different viewpoints against a static background and with good illumination. The 3D proposal grid consists of $10^3$ voxels, with cuboid side length of 4 meters.

\parag{Metrics.} We report our segmentation scores in J- and F-measure as defined in~\cite{Perazzi16}. The former is defined as the intersection-over-union (IoU) between the ground-truth segmentation mask and the prediction, while the latter is the harmonic average between the precision and the recall at the mask boundaries. The detection scores are calculated in terms of mAP$_{0.5}$, the mean probability of having an IoU of more than 50$\%$.  Different segmentation algorithms set the foreground-background threshold differently. Hence, to allow a fair comparison, we perform a line search from 0 to 1 with a step-size of 0.05, selecting the optimal value for all baselines and variants for each individual dataset. 

\subsection{Comparative Results with Moving Cameras}


\renewcommand{\arraystretch}{1}
\renewcommand{\tabcolsep}{2mm}
\begin{table}[t]
	\small
	\begin{center}
		\resizebox{1.0\columnwidth}{!}{
			
			\begin{tabular}{@{}lccc@{}}
				&\multicolumn{3}{c}{\ski}\\
				\toprule
				Method              &J Score &F Score  &Run-time (sec)\\
				\midrule
				Chen et al.~\cite{Chen19a}			&0.37				&0.42    &0.11 \\
				Stretcu et al.~\cite{Stretcu15}		&0.51			&0.56		&0.02	\\
				Lu et al.~\cite{Lu20}						&0.51					&0.60					& 0.60        \\
				Katircioglu et al.	\cite{Katircioglu20}			&0.61		&0.67	&0.24	\\

				Rhodin et al.~\cite{Rhodin19a} + \cite{Katircioglu20} 		&0.61		&0.70	&0.23\\
				Croitoru et al. \cite{Croitoru19}		& 0.62	 &0.72	 &0.15\\
				Yang et al. \cite{Yang19c} w/o CRF			& 0.61 	 &0.71  &0.32\\
				Yang et al. \cite{Yang19c}			& 0.67 	 &0.77  &1.12\\
				Katircioglu et al.	\cite{Katircioglu20} w/ flow			&0.69		&0.79		&0.24\\
				Koh et al. \cite{Koh17b}				&0.70 	 &0.80		&107.4\\
				Ours				& \bf{0.71} 		&\bf{0.83}		&0.17\\
				\bottomrule
			\end{tabular}
		}
	\end{center}
	\vspace{-6mm}
	\caption{\textbf{Segmentation results on the \ski{}.} We compare against the state-of-the-art single-view approaches and a modified version of the multi-view approach of~\cite{Rhodin19a}.} 	\label{tbl:results_ski_quantitative}
	\vspace{-5mm}
\end{table}


\begin{figure*}[th!]
	\centering
	{
		\resizebox{1.0\linewidth}{!}{
			\begin{tabular}{cccccc}
				\includegraphics[width=0.80\textwidth]{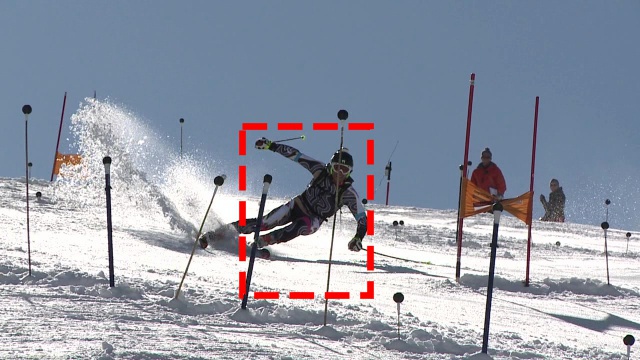}&
				\includegraphics[width=0.80\textwidth]{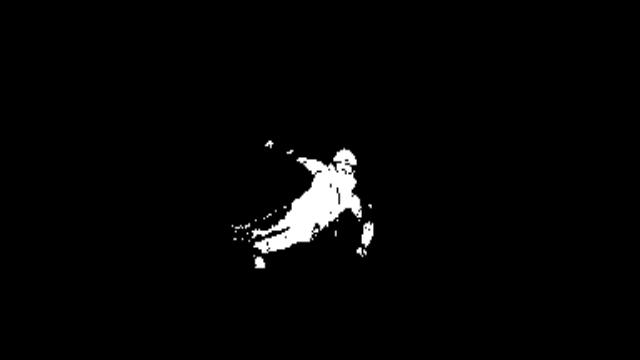}&
				
				\includegraphics[width=0.80\textwidth]{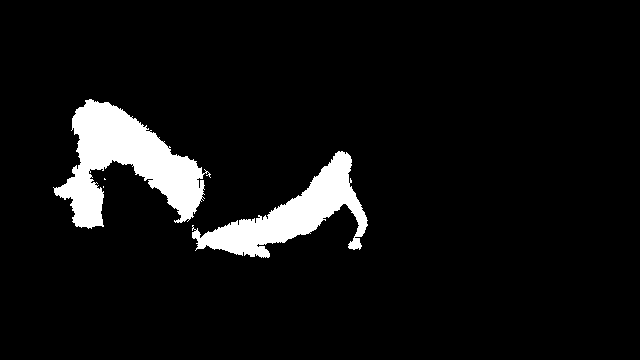} &
				\includegraphics[width=0.80\textwidth]{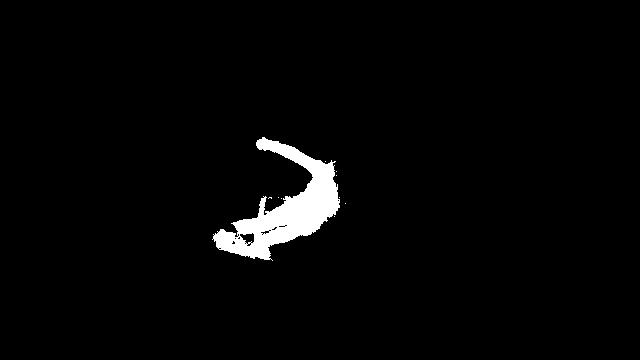} &
				\includegraphics[width=0.80\textwidth]{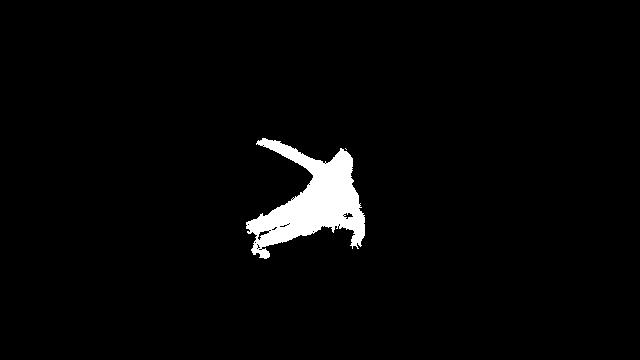} &
				\includegraphics[width=0.80\textwidth]{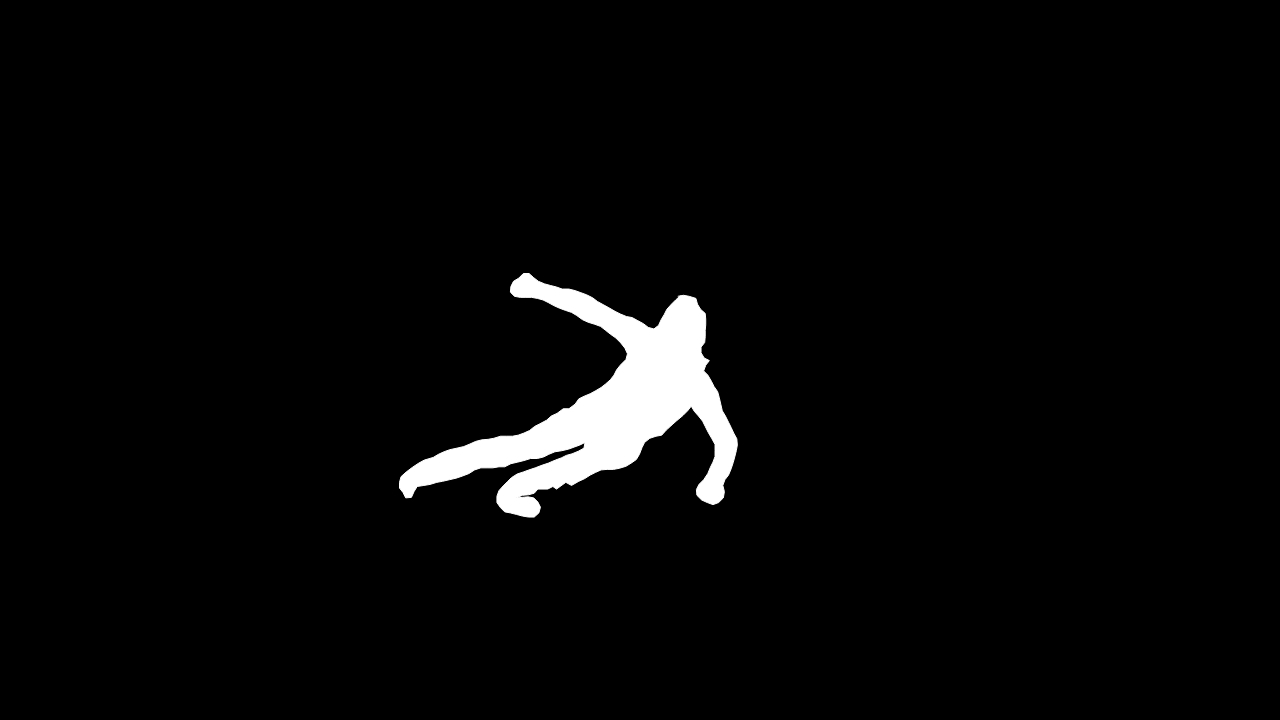} \\

				\includegraphics[width=0.80\textwidth]{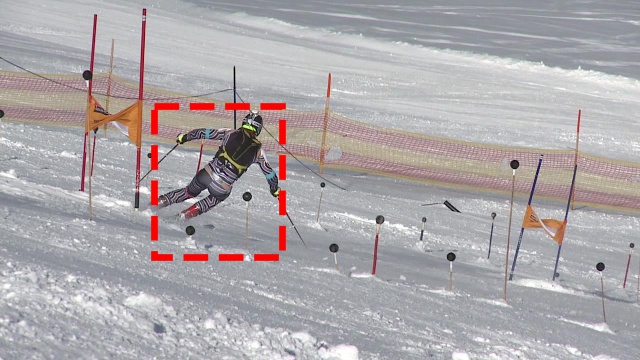} &
				\includegraphics[width=0.80\textwidth]{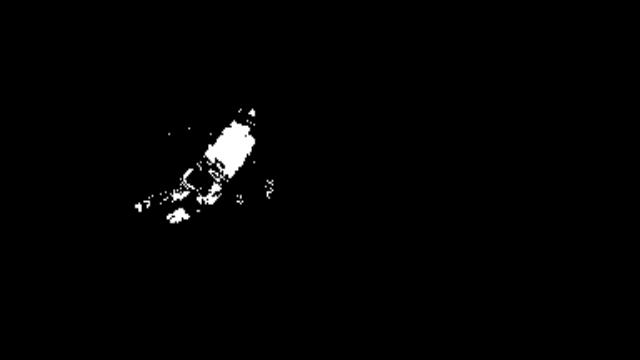} &
				\includegraphics[width=0.80\textwidth]{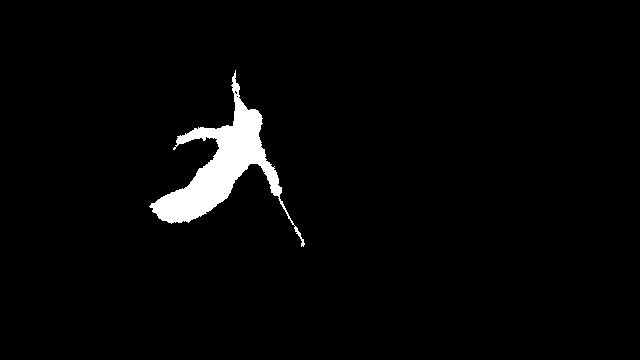} &
				\includegraphics[width=0.80\textwidth]{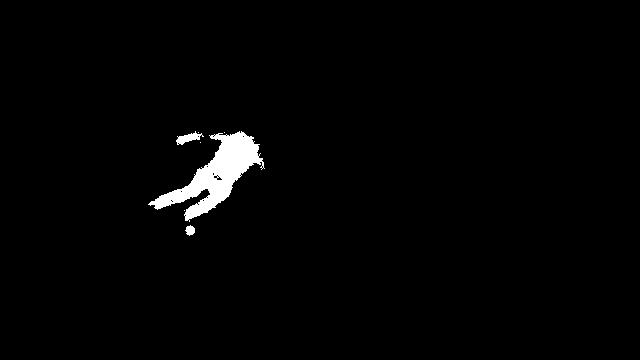} &
				\includegraphics[width=0.80\textwidth]{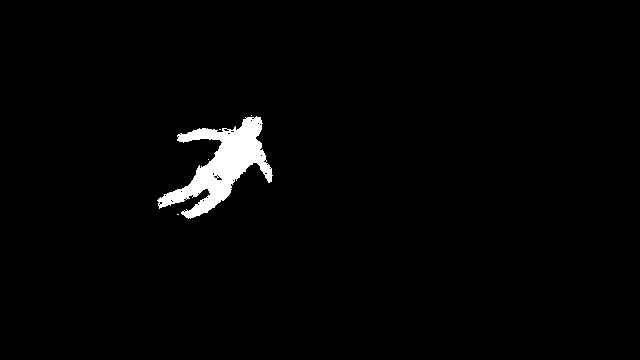} &
				\includegraphics[width=0.80\textwidth]{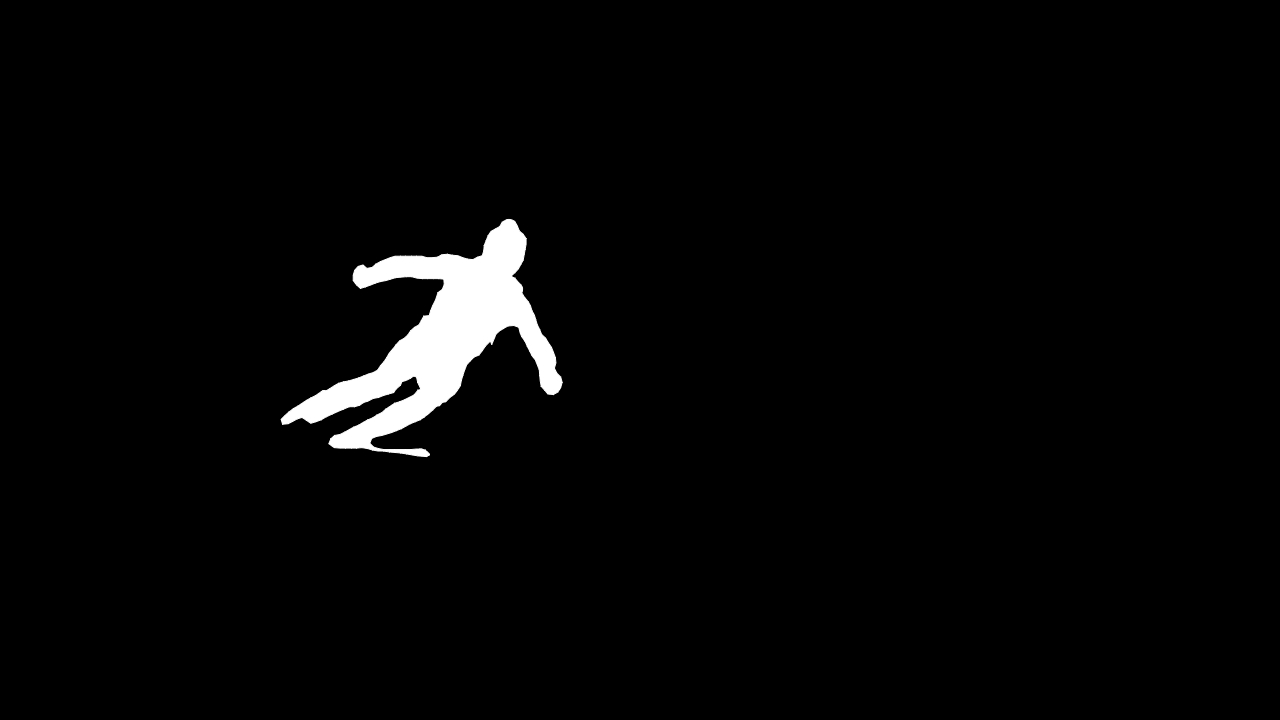} \\

				\includegraphics[width=0.80\textwidth]{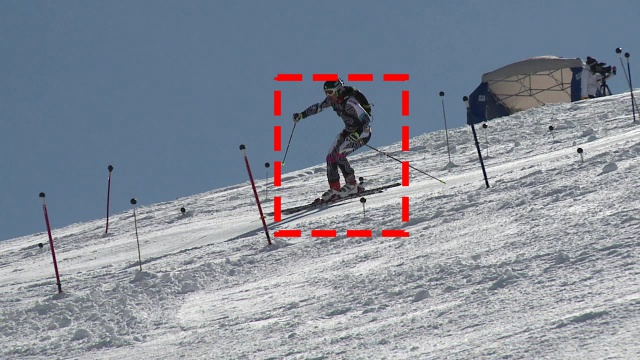} &
				\includegraphics[width=0.80\textwidth]{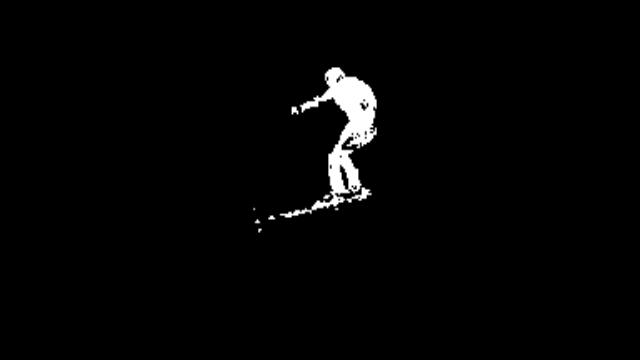} &
				\includegraphics[width=0.80\textwidth]{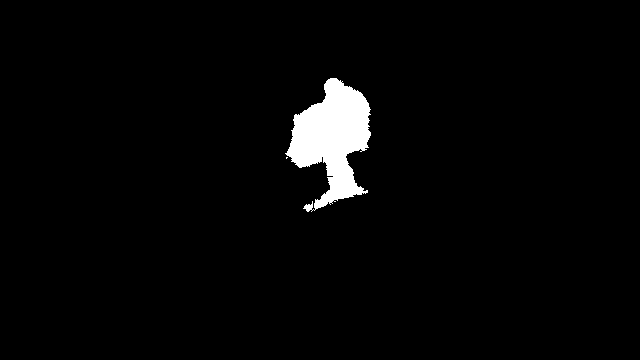} &
				\includegraphics[width=0.80\textwidth]{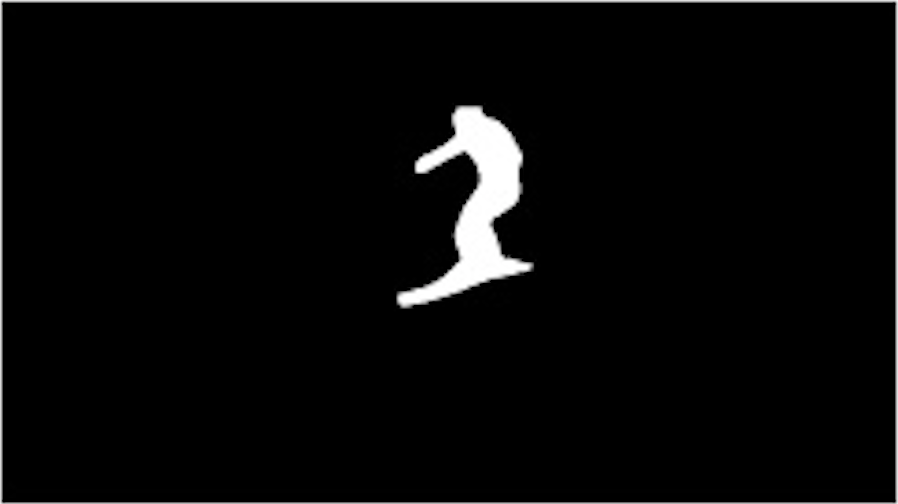} &
				\includegraphics[width=0.80\textwidth]{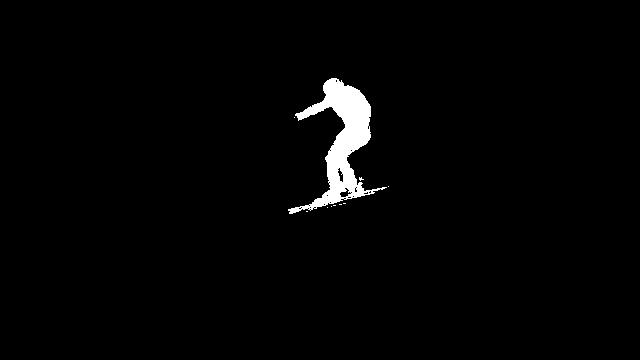} &
				\includegraphics[width=0.80\textwidth]{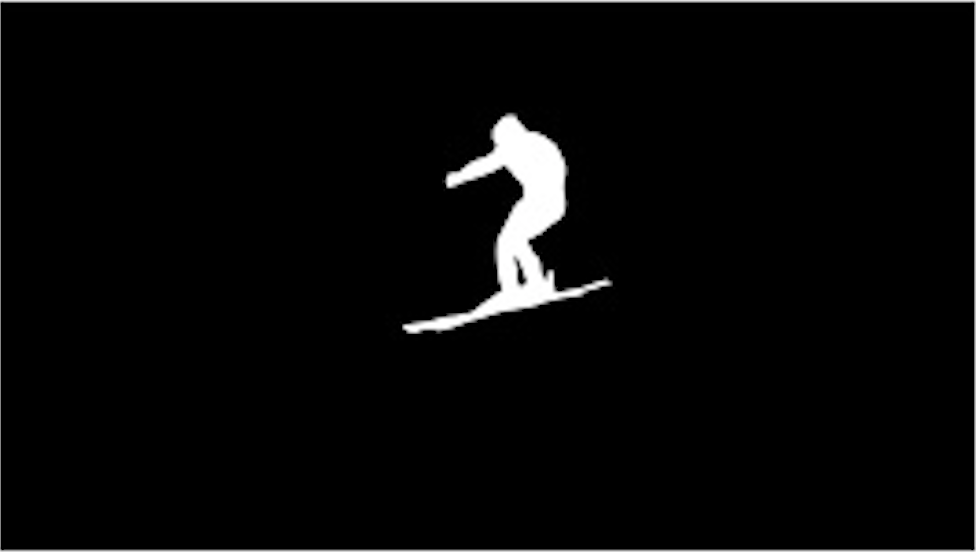} \\ \\

				{\fontsize{40}{5}\selectfont  (a)Input/Ours detection} &  {\fontsize{40}{5}\selectfont  (b) Yang et al.~\cite{Yang19c}}  & {\fontsize{40}{10}\selectfont  (c) Koh et al.~\cite{Koh17b}} & {\fontsize{40}{10}\selectfont  (d) Katircioglu et al.~\cite{Katircioglu20}} & {\fontsize{40}{10}\selectfont  (e) Ours} & {\fontsize{40}{10}\selectfont  (f) GT }   \\ 
		\end{tabular}}
	}
	\vspace{-2mm}
	\caption{\textbf{Qualitative results on the \ski{} dataset.} (a) Input images with our predicted bounding box overlaid in red. (b,c,d) Segmentation masks predicted by three of our baselines. (e) Our segmentation mask prediction. (f) Ground truth segmentation mask. Note the quality of our predicted masks even though, unlike the methods of~\cite{Koh17b} and~\cite{Yang19c}, we do not use explicit temporal cues at inference time.}
	\label{fig:ski_qual}
		\vspace{-4mm}
\end{figure*}

Fig.~\ref{fig:ski_qual} depicts qualitative results on the \ski{} dataset and we report the corresponding quantitative results using 4 cameras in Table~\ref{tbl:results_ski_quantitative}, in which we use the scores reported in~\cite{Katircioglu20} for the baselines. \footnote{The implementations of~\cite{Katircioglu20} and~\cite{Lu20} were provided by the authors.}


We outperform all existing single-view self-supervised segmentation approaches~\cite{Koh17b,Yang19c,Croitoru19,Katircioglu20,Stretcu15,Chen19a,Lu20} while being comparatively fast. For completeness, we also report results for~\cite{Yang19c} without CRF post-processing. This shows that a great deal of the method's performance comes from such post-processing, which we do not require. Note that, in contrast to~\cite{Katircioglu20} with flow and~\cite{Koh17b}, our approach does not require computing optical flow. Unlike DAVIS~\cite{Perazzi16}, our datasets feature large camera motions with quick background changes, which causes methods such as~\cite{Lu20} to often merge portions of the background and the human.


The only other self-supervised multi-view approach for which a public implementation is available is that of~\cite{Rhodin19a}. Unfortunately, it requires background images as an input, which are not given in this case and are not trivial to create because the cameras rotate and zoom. To do so anyway, we use the single-view approach of~\cite{Katircioglu20} to produce background images that we can feed to the network of~\cite{Rhodin19a} for multi-view training. As can be seen in Table~\ref{tbl:results_ski_quantitative}, this modified version of~\cite{Rhodin19a} does slightly better than~\cite{Katircioglu20} in F score terms but remains far behind our method. The method that comes closest to ours is that of~\cite{Koh17b}, which operates on the whole sequence and is therefore prohibitively slow as discussed below.  By contrast our approach operates on a single image and does not require motion information.

The inference times for each method are shown in the last column of Table~\ref{tbl:results_ski_quantitative} and computed using code that is either publicly available or that the authors made available to us privately. All except those of~\cite{Stretcu15,Koh17b} were obtained using a single NVIDIA TITAN X Pascal GPU. Since~\cite{Stretcu15,Koh17b} are designed to run on CPU, the inference for them is computed on Intel(R) Xeon(R) Gold 6240 CPUs. The tailored optimization approach of~\cite{Koh17b} that comes closest to our results is three orders of magnitude slower than our approach because it tracks several patches over time. Unlike~\cite{Yang19c}, our method does not require optical flow computation or CRF post-processing which brings a five-fold speedup. Our computational complexity is similar to that of~\cite{Katircioglu20, Rhodin19a} since the triangulation time is negligible. The training time of our model on the \ski{} is approximately $8$ hours whereas that of~\cite{Rhodin19a} and~\cite{Katircioglu20} are $14$ and $7.5$ hours, respectively. 



\begin{figure}[t]
  \centering
  {
  	\resizebox{\linewidth}{!}{
\renewcommand{\arraystretch}{0.7}
  \begin{tabular}{@{}ccccc@{}}
   \includegraphics[width=1.1\linewidth]{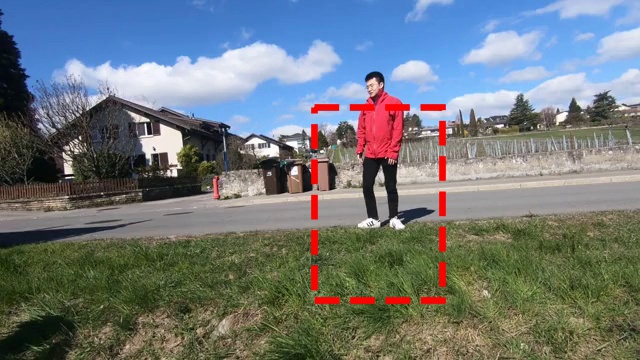}
   \includegraphics[width=1.1\linewidth]{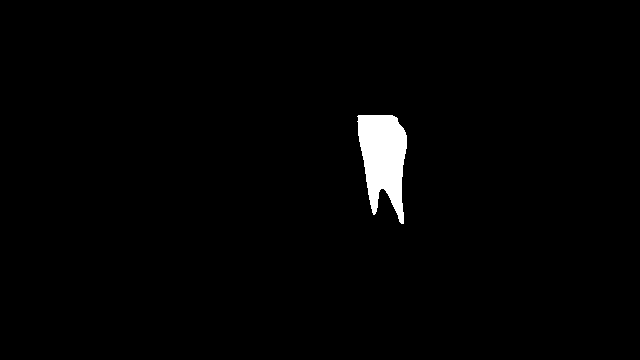}&
   
      \includegraphics[width=1.1\linewidth]{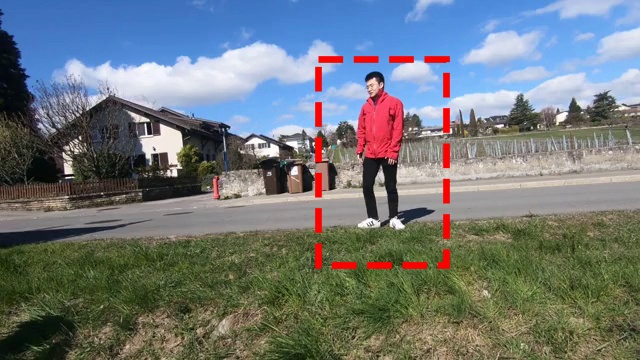} 
   \includegraphics[width=1.1\linewidth]{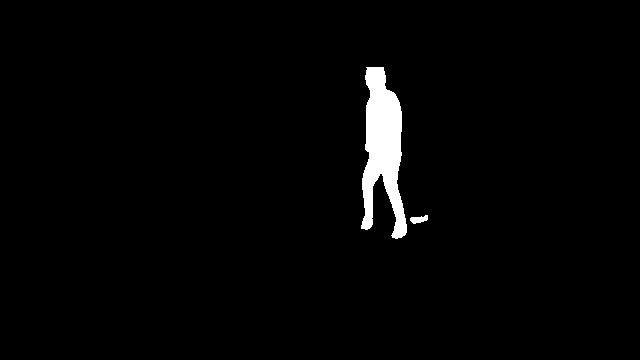} &
   
   \includegraphics[width=1.1\linewidth]{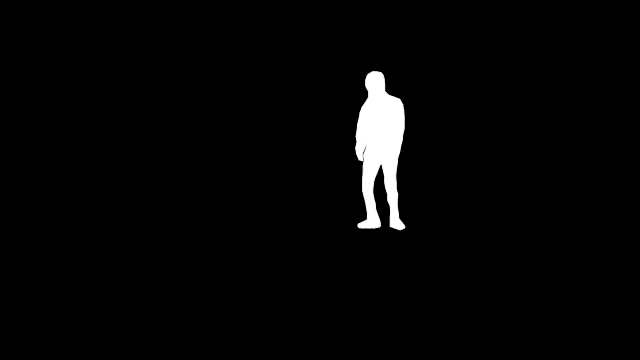} \\

     \includegraphics[width=1.1\linewidth]{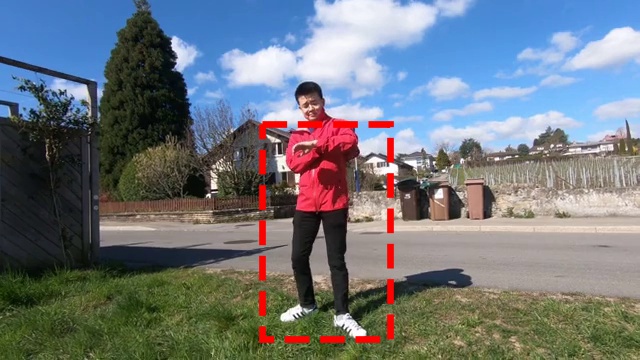}
   \includegraphics[width=1.1\linewidth]{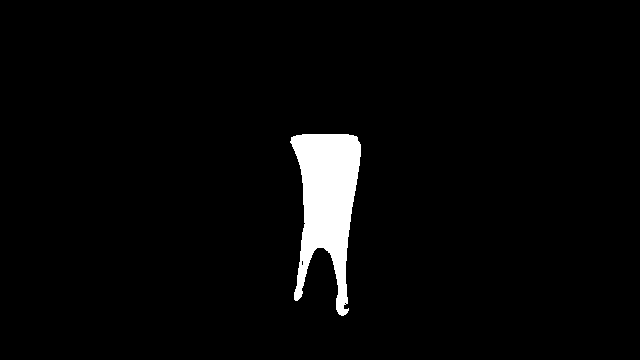}&
   
   \includegraphics[width=1.1\linewidth]{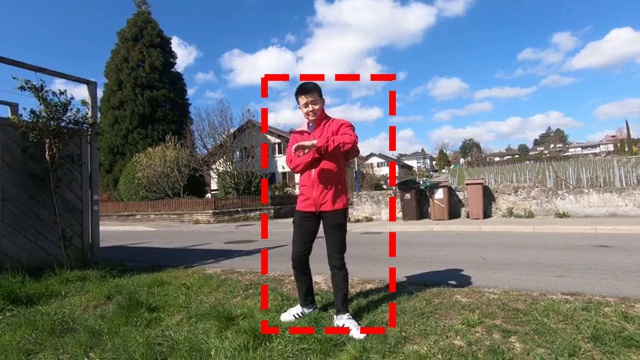} 
   \includegraphics[width=1.1\linewidth]{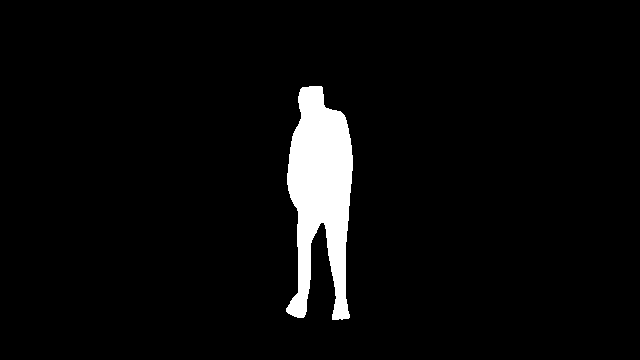} &
   
   \includegraphics[width=1.1\linewidth]{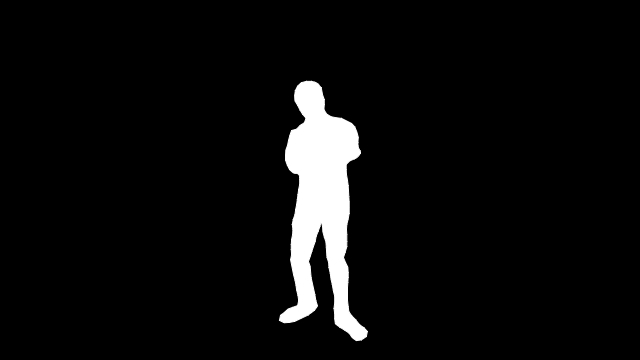} \\ 
   
    \includegraphics[width=1.1\linewidth]{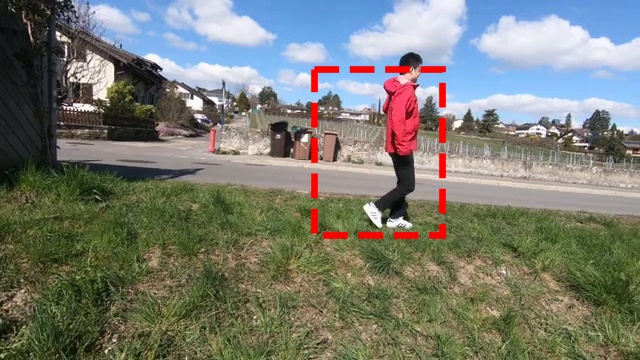}
   \includegraphics[width=1.1\linewidth]{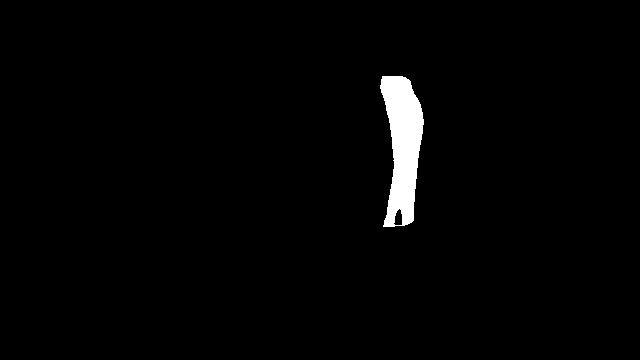}&
   
   \includegraphics[width=1.1\linewidth]{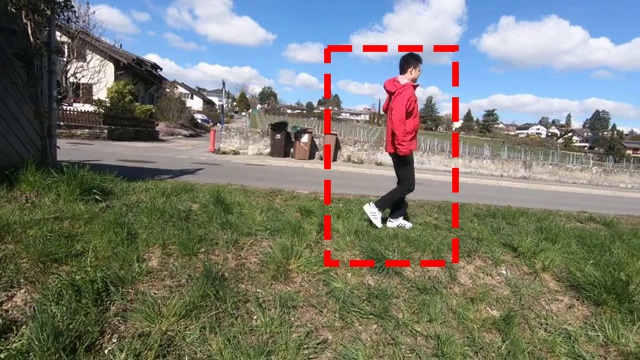} 
   \includegraphics[width=1.1\linewidth]{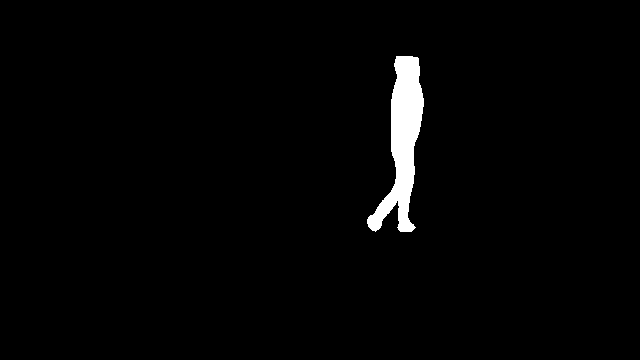} &
   
   \includegraphics[width=1.1\linewidth]{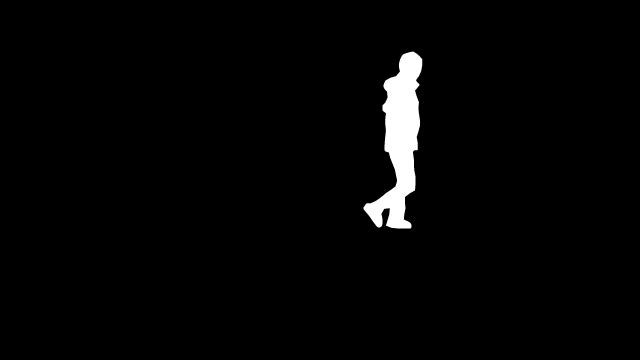} \\ \\

{\fontsize{40}{5}\selectfont   (a)~Katircioglu et al.~\cite{Katircioglu20}} &  {\fontsize{40}{5}\selectfont  (b) Ours}  & {\fontsize{40}{5}\selectfont  (c) GT }  \\ 
  \end{tabular}}
}
\vspace{-2mm}
  \caption{ \textbf{Qualitative results on the {\handheld{}} dataset.} (a) The detection and segmentation mask results of~\cite{Katircioglu20} trained and tested on single images. (b) The predictions of our model trained using 3-camera multi-view consistency and tested on single images. (c) Ground truth. Our results are generally more accurate, which justifies the effort invested in calibrating the cameras.}
  \label{fig:handheld_qual}
  \vspace{-4mm}
\end{figure}

We also evaluate our method on \handheld{} trained using 4200 images from multiple views and compare against the network of~\cite{Katircioglu20} trained using the same 4200 images. We obtain a J-score of $0.66$ instead of $0.64$ and an F-score of $0.77$ instead of $0.71$, again showing the importance of multi-view consistency.
Our method benefits from multi-view information obtained in an automated off-the-shelf manner, particularly in tightly fitting to the subject, as shown in Fig.~\ref{fig:handheld_qual}. In short, the improvement demonstrated here highlights the previously untapped potential of multi-view constraints for self-supervised segmentation.


\renewcommand{\arraystretch}{1}
\renewcommand{\tabcolsep}{2mm}
\begin{table}[t]
	\small
	\begin{center}
		\resizebox{1\columnwidth}{!}{
			
			\begin{tabular}{@{}lccc@{}}
				& &\multicolumn{2}{c}{\human}\\
				\toprule
				Method    & Training Type & Background Assumption  &     mAP \\
				\midrule
				Katircioglu et al.~\cite{Katircioglu20}	& single-view & dynamic	& 0.57 		\\
				Rhodin et al.~\cite{Rhodin19a} & multi-view &static & 0.71 			\\
				Ours						& multi-view & dynamic & \bf{0.85} 				\\
				\bottomrule
			\end{tabular}
		}
	\end{center}
	\vspace{-6mm}
	\caption{ \textbf{Comparative results on the \human{} dataset.} Our detection accuracy improves in terms of mAP$_{0.5}$. }
	\label{table:results_h36m}
\end{table}


\begin{figure}[t]
  \centering
  {
  	\resizebox{\linewidth}{!}{
\renewcommand{\arraystretch}{0.5}
  \begin{tabular}{ccccccc}
   \includegraphics[width=1.0\linewidth]{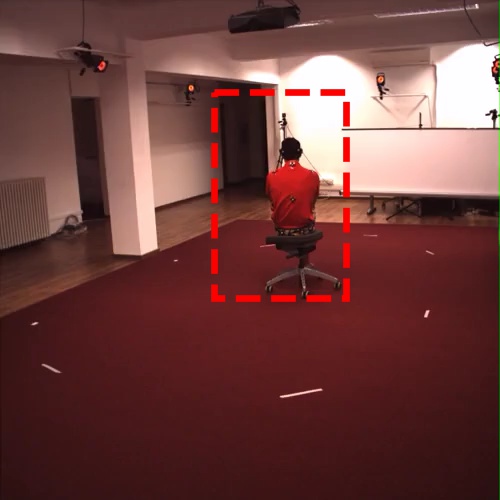} 
   \includegraphics[width=1.0\linewidth]{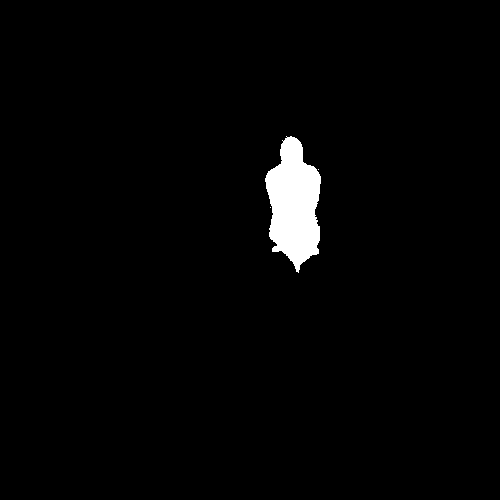}&
   \hspace{-5mm}
      \includegraphics[width=1.0\linewidth]{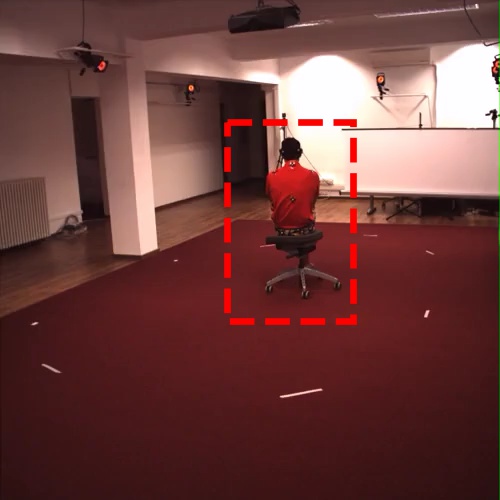} 
   \includegraphics[width=1.0\linewidth]{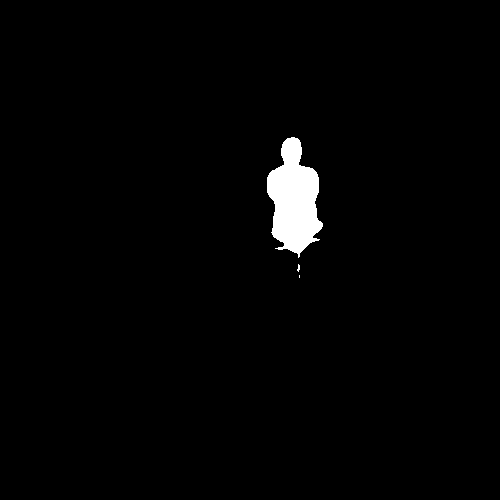} &
      \includegraphics[width=1.0\linewidth]{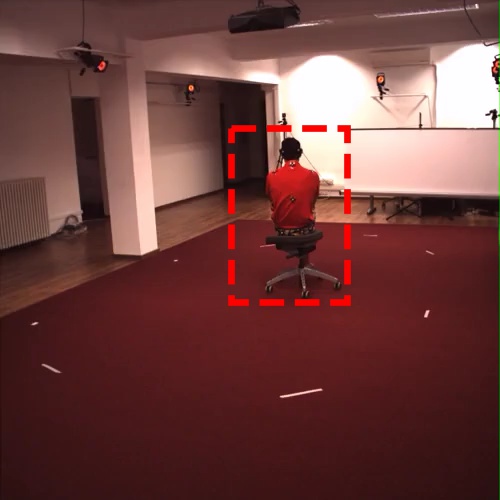} 
   \includegraphics[width=1.0\linewidth]{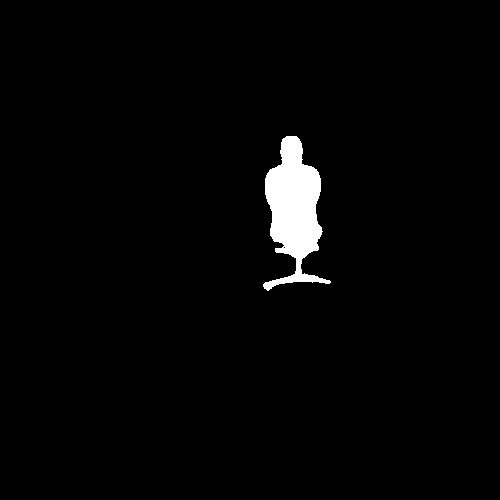} &
   \includegraphics[width=1.0\linewidth]{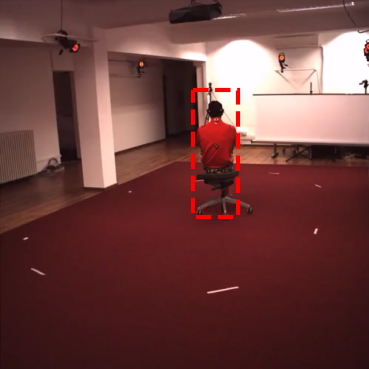} \\

     \includegraphics[width=1.0\linewidth]{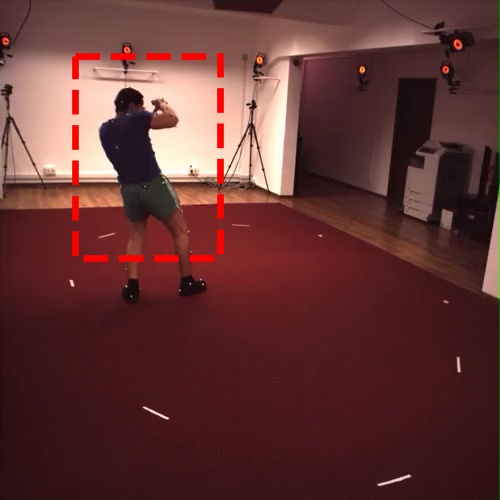}
   \includegraphics[width=1.0\linewidth]{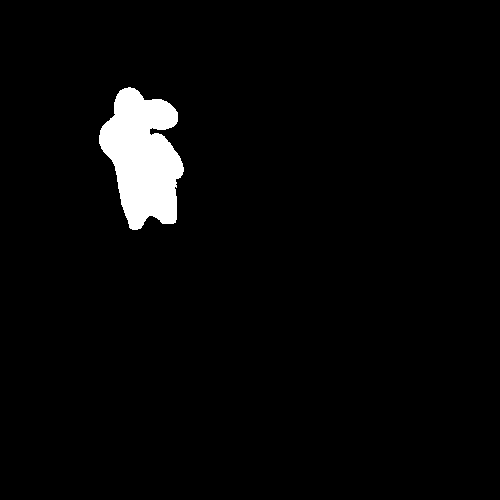}&
      \hspace{-5mm}
   \includegraphics[width=1.0\linewidth]{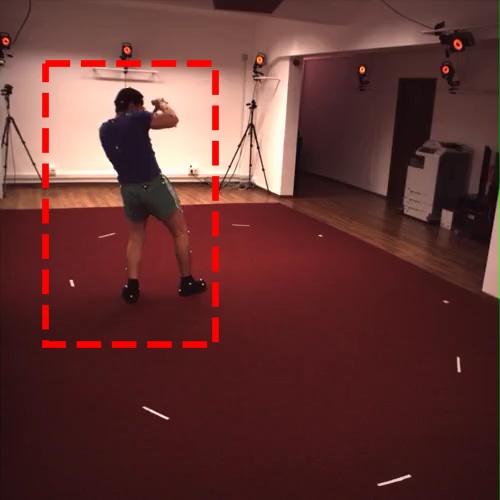} 
   \includegraphics[width=1.0\linewidth]{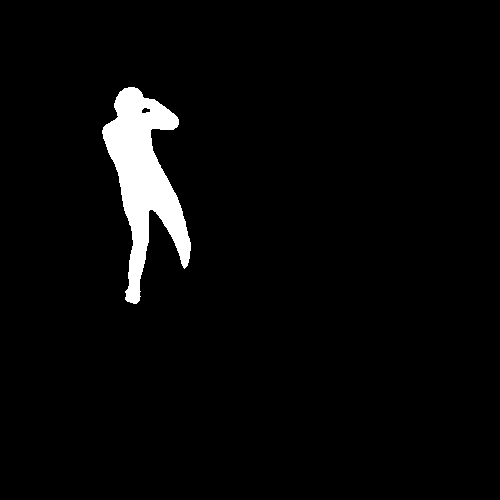} &
   \includegraphics[width=1.0\linewidth]{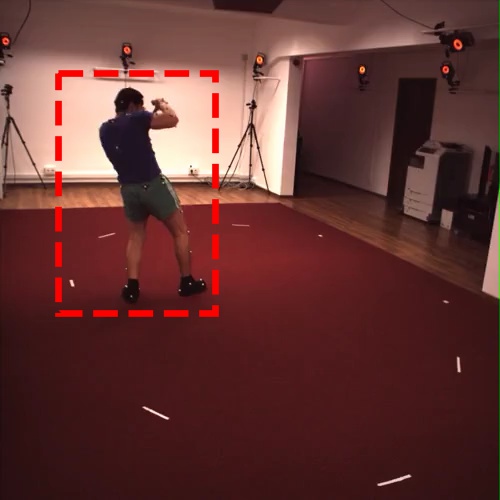} 
   \includegraphics[width=1.0\linewidth]{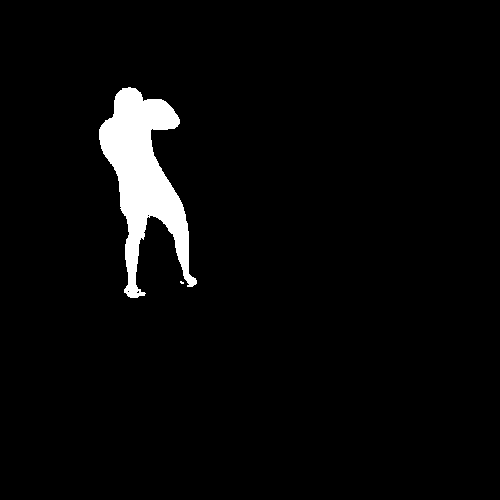} &
   \includegraphics[width=1.0\linewidth]{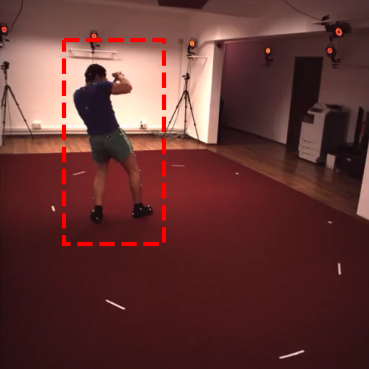} \\

 \includegraphics[width=1.0\linewidth]{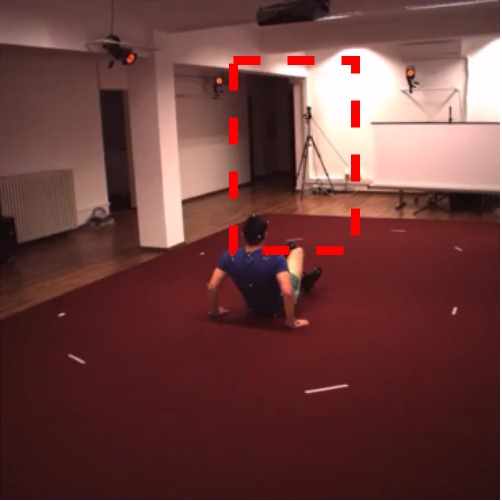}
\includegraphics[width=1.0\linewidth]{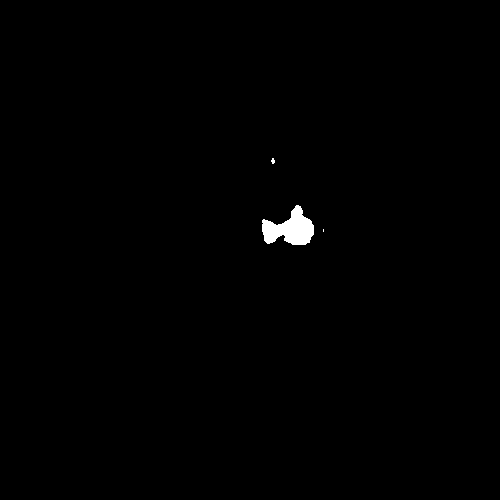}&
   \hspace{-5mm}
\includegraphics[width=1.0\linewidth]{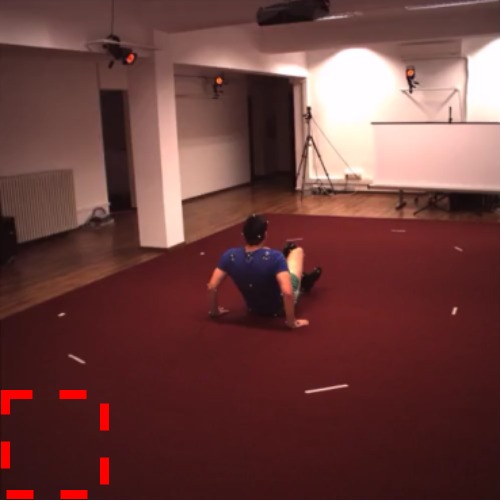} 
\includegraphics[width=1.0\linewidth]{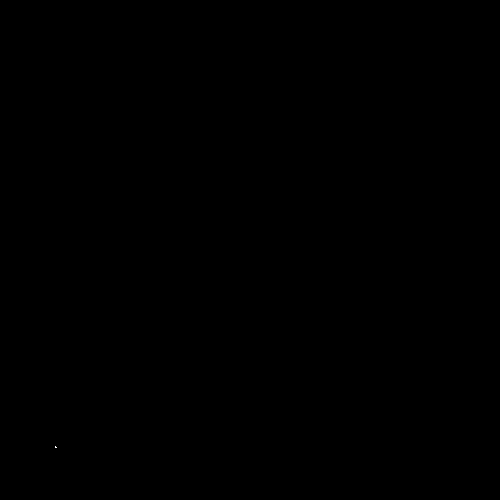} &
\includegraphics[width=1.0\linewidth]{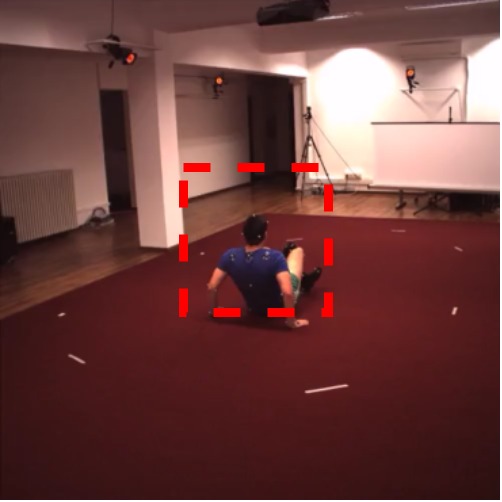} 
\includegraphics[width=1.0\linewidth]{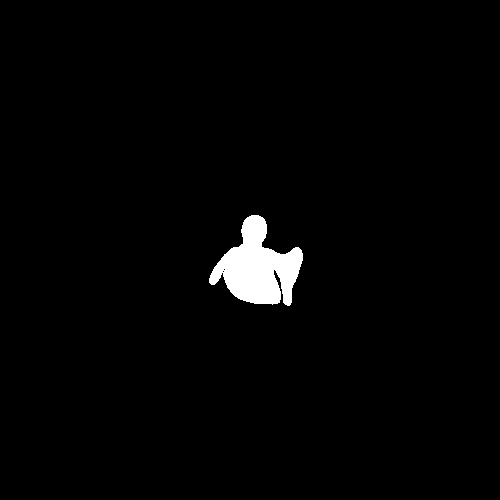} &
\includegraphics[width=1.0\linewidth]{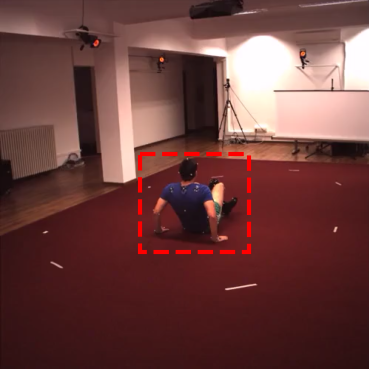} \\ \\ \\

\hspace{-5mm}{\fontsize{51}{1}\selectfont  (a)~Katircioglu et al.~\cite{Katircioglu20}}& {\fontsize{51}{1}\selectfont (b) Rhodin et al.~\cite{Rhodin19a}} &  {\fontsize{51}{1}\selectfont (c) Ours}  &  {\fontsize{51}{1}\selectfont (d) GT }  \\ 
  \end{tabular}}
}
\vspace{-2mm}
  \caption{ \textbf{Qualitative results on the {\human} dataset.} (a) The detection and segmentation results of~\cite{Katircioglu20} trained and tested on single images. (b) The results of~\cite{Rhodin19a} trained with a pair of camera views and tested on single images. (c) Our predictions obtained from the model trained with the 4-cam multi-view consistency and tested on single images. (d) Ground truth. Our method consistently detects the person whereas~\cite{Katircioglu20,Rhodin19a} occasionally produce inconsistent results, such as the failed detections in the last row.}
  \label{fig:h36m_qual}
  \vspace{-4mm}
\end{figure}

\subsection{Comparative Results with Static Cameras}

In the previous example, we had to modify the multi-view self-supervised algorithm of~\cite{Rhodin19a} to make it work on images with a moving background.  To evaluate the original version instead, we compare on the \human{} dataset and report the results using again 4 cameras in Table~\ref{table:results_h36m}. As in the \ski{} case, we outperform it and, this time, the difference cannot be caused by any background modification we made. This is somewhat surprising because the method of~\cite{Rhodin19a} assumes a constant static background, which is the case here, whereas ours is learned without any such constraint. We attribute this result to the explicit consistency of bounding box positions in 3D and the background inpainting constraint. The latter triggers when part of the subject is outside the bounding box leading to correctly segmented legs while the method of~\cite{Rhodin19a} has trouble distinguishing the skin and floor color when in shadow, as depicted in Fig.~\ref{fig:h36m_qual}. See additional qualitative results in the supplementary.

In Table~\ref{table:results_h36m}, we also report the result of~\cite{Katircioglu20}, that is, our backbone network run on single views. The performance drops, which once again highlights the usefulness to exploit multiple views for training when they are available.

\begin{table}[t] \centering
	\resizebox{\linewidth}{!}{%
	\begin{tabular}{ c*{6}c @{}}
				\cmidrule{1-7} 
		& {\# Cam}  & \textit{Ours w/o VC} & \textit{Ours w/o HC} & \textit{Ours w/ TC}
		& \textit{Ours w/ WC} & \textit{Ours}\\
		\cmidrule{1-7} 
		& 2              & 0.66 & 0.67 & 0.66 &0.61 & 0.66 \\
		& 3               &0.68   & 0.70  &0.68   &0.68 & \bf{0.71} \\
		& 4  &0.68 &0.70  & 0.67 &0.68 &\bf{0.71}  \\
				\rot{\rlap{~J Score}} 
		& 5  & 0.67 & 0.67 & 0.67 &0.68 & 0.69  \\
		& 6  & 0.66 & 0.70 &0.67  &0.67 &0.68  \\

		\cmidrule{1-7}

	& 2              & 0.73 & 0.73  & 0.73 &0.65 & 0.75 \\
& 3               & 0.75 & 0.77  & 0.75 &0.74 & 0.81\\
& 4  & 0.75 &{0.79}  & 0.75 &0.77 & \bf{0.83} \\
\rot{\rlap{~F Score}} 
& 5  & 0.74 & 0.74  & 0.74 &0.75 & 0.78  \\
& 6  & 0.73 & 0.78  & 0.73 &0.74 & 0.76\\
	\cmidrule{1-7}
\end{tabular}}
	\vspace{-4mm}
	\caption{ \textbf{Ablation study on the {\ski}}. We test variants of our approach while using varying numbers of cameras.}
	\vspace{-5mm}
	\label{table:ski_ablation}
\end{table}


\begin{table}[t] \centering
	\resizebox{\linewidth}{!}{%
			\begin{tabular}{ c*{6}c @{}}
				\cmidrule{1-7} 
		& {\# Cam}  & \textit{Ours w/o VC} & \textit{Ours w/o HC} & \textit{Ours w/ TC}
		& \textit{Ours w/ WC} & \textit{Ours}\\
		\cmidrule{1-7} 
		& 2              & 0.73 & 0.74 & 0.74 & 0.73 & 0.75\\
		
		& 3               & 0.78  & 0.80   & 0.79  & 0.79 & 0.82 \\
			\rot{\rlap{~mAP}} 
		& 4  & 0.79 & 0.83 & 0.82 & 0.84 & \bf{0.85}\\

		\cmidrule{1-7}
		
\end{tabular}
}
         \vspace{-4mm}
	\caption{ \textbf{Ablation study on the {\human} dataset.} We test variants of our approach while using varying numbers of cameras.} 
	\label{table:h36m_ablation}
	\vspace{-5mm}
\end{table}

\subsection{Ablation Study}

We compare the following variants of the multi-view constraints of Section~\ref{multi-view training}:
\textit{Ours} denotes the full model that employs all the steps shown in Fig.~\ref{fig:multiview}.
\textit{Ours w/o HC} excludes the bounding box height consistency depicted by Fig.~\ref{fig:multiview} (c). 
\textit{Ours w/o VC} leaves out both the center and height adjustment of Fig.~\ref{fig:multiview} (b,c) and enforces only consistent sampling. 
\textit{Ours w/ WC} imposes a bounding box width consistency in addition to the  full model. 
Finally, \textit{Ours  w/ TC} is a baseline that replaces the view consistency with a triangulation loss minimizing the distance between the lines joining the centers of the camera and predicted 2D bounding box.

In Table~\ref{table:ski_ablation} and Table~\ref{table:h36m_ablation} we report results as a function of the number of cameras we used. We can use only 2 cameras but the best results are obtained for 3 or 4. Beyond that, additional cameras add little new information while taking more space in the training batches, resulting in less diverse batches and lower performance. The numbers for the different variants in Fig.~\ref{fig:multiview} show that all the elements we have incorporated into our approach contribute positively and that the one we have purposely ignored---constraining the width---would degrade performance. Crucially  \textit{Ours  w/ TC} also performs worse, hence substantiating our claim that imposing consistency constraints using the projection mechanism of Section~\ref{sec:consistency} is crucial to our success.

We also analyzed the influence of the voxel resolution on the reconstruction accuracy. Table~\ref{table:voxel_ablation} shows that a $10^3$ cube is more accurate than a $6^3$ cube while going to a $16^3$ does not bring further improvements in \human{} dataset. The 0.01 lower mAP may indicate that learning a discrete distribution on the 3D grid may be less efficient on larger spaces. However, as the ski footage covers a wider area, a $16^3$ cube yields the best performance on the \ski{}.


\renewcommand{\arraystretch}{1}
\renewcommand{\tabcolsep}{2mm}
\begin{table}
	\centering
	\resizebox{0.60\textwidth}{!}{
		\hspace{-25mm}
		\begin{minipage}{0.55\textwidth}
			\centering
			\begin{tabular}{@{}lc@{}}
				3D Grid Size &\multicolumn{1}{c}{\textbf{Ski-PTZ} \footnotesize{J-Score}}\\
				\toprule
				${[10 \times 10 \times 10]}$ &0.64 \\
				${[16 \times 16 \times 16]}$& \bf{0.68}	\\
				${[24 \times 24 \times 24]}$	&0.66	\\
				\bottomrule
			\end{tabular}
		\end{minipage}
		\hspace{-48mm}
		\begin{minipage}{0.55\textwidth}
			\centering
			\begin{tabular}{@{}lc@{}}
				3D Grid Size & \multicolumn{1}{c}{\human} \footnotesize{mAP$_{0.5}$}\\
				\toprule
				${[6 \times 6 \times 6]}$ &0.76\\
				${[10 \times 10 \times 10]}$& \bf{0.79}	\\
				${[16 \times 16 \times 16]}$	&0.76	\\
				\bottomrule
			\end{tabular}
	\end{minipage}} \quad\quad
	\vspace{-3mm}
	\caption{\textbf{Influence of voxel resolution.} The numbers in square brackets indicate the number of voxels in the 3D proposal grid and we use $4$ cameras.}
	\label{table:voxel_ablation}
	\vspace{-6mm}
\end{table}


\vspace{-2mm}
\section{Conclusion}
We have presented a self-supervised detection and segmentation technique that exploits multi-view geometry during training to accurately separate foreground from background in single RGB images at inference time. It outperforms the state-of-the-art on the challenging {\ski}, depicting unusual activities captured with moving cameras, and on {\human}, acquired with static cameras. We have 
focused on scenes with a single salient object. However, our method has the potential to handle multiple objects by sampling more than one proposal as long as they are not overlapping. Our future work will be in this direction.


{\small
	\bibliographystyle{ieee_fullname}
	\bibliography{string,graphics,vision,learning,misc}
}

\end{document}